\documentclass[10pt,twocolumn,letterpaper,pagebackref,breaklinks,colorlinks,allcolors=cvprblue]{article}

\usepackage[pagenumbers]{cvpr}              
\definecolor{cvprblue}{rgb}{0.21,0.49,0.74}

\usepackage{hyperref}
\usepackage{adjustbox}
\usepackage{tabularx}
\usepackage{makecell}
\usepackage{subcaption}
\usepackage{pifont}
\usepackage{stmaryrd}
\usepackage{subcaption}
\usepackage{makecell}
\usepackage{adjustbox}
\usepackage{longtable}  
\usepackage{booktabs}   
\usepackage{mathastext}
\MTfamily{dejavusans}
\usepackage{graphicx}%
\usepackage{multirow}%
\usepackage{amsmath,amssymb,amsfonts}%
\usepackage{amsthm}%
\usepackage{mathrsfs}%
\usepackage[title]{appendix}%
\usepackage{pifont}%
\usepackage{textcomp}%
\usepackage{manyfoot}%
\usepackage{booktabs}%
\usepackage{algorithm}%
\usepackage{algorithmicx}%
\usepackage{algpseudocode}%
\usepackage{listings}%
\usepackage{mathpazo} 
\usepackage[justification=raggedright,singlelinecheck=false]{caption}

\title{SurgVLM: A Large Vision-Language Model and Systematic Evaluation Benchmark for Surgical Intelligence}

\author{Zhitao Zeng$^{1,\dagger}$
\and
Zhu Zhuo$^{1,\dagger}$
\and
Xiaojun Jia$^{2,\dagger}$
\and
Erli Zhang$^{1,\dagger}$
\and
Junde Wu$^{3,\dagger}$
\and
Jiaan Zhang$^1$
\and
Yuxuan Wang$^{4}$
\and
Chang Han Low$^{1}$
\and
Jian Jiang$^{5}$
\and
Zilong Zheng$^{4}$
\and
Xiaochun Cao$^6$
\and
Yutong Ban$^5$
\and
Qi Dou$^7$
\and
Yang Liu$^2$ 
\and
Yueming Jin$^{1,^{\ast}}$ \\$ $ \\ $ $ \\ 
$^1$National University of Singapore \\$^2$Nanyang Technological University \hspace{1em} $^3$University of Oxford \\ $^4$State Key Laboratory of General Artificial Intelligence, BIGAI \\$^5$Shanghai Jiao Tong University \hspace{1em} $^6$Sun Yat-sen University \\$^7$The Chinese University of Hong Kong
\\ $ $
\href{https://jinlab-imvr.github.io/SurgVLM}{https://jinlab-imvr.github.io/SurgVLM}
}

\begin{document}
\twocolumn[{%
\maketitle
\renewcommand\twocolumn[1][]{#1}%
\vspace{-1em}

\begin{abstract}

Foundation models have achieved transformative success across biomedical domains by enabling holistic understanding of multimodal data. However, their application in surgery remains underexplored. Surgical intelligence presents unique challenges - requiring surgical visual perception, temporal analysis, and reasoning. Existing general-purpose vision-language models fail to address these needs due to insufficient domain-specific supervision and the lack of a large-scale high-quality surgical database. To bridge this gap, we propose SurgVLM, one of the first large vision-language foundation models for surgical intelligence, where this single universal model can tackle versatile surgical tasks.
To enable this, we construct a large-scale multimodal surgical database, SurgVLM-DB, comprising over 1.81 million frames with 7.79 million conversations, spanning more than 16 surgical types and 18 anatomical structures. 
We unify and reorganize 23 public datasets across 10 surgical tasks, followed by standardizing labels and doing hierarchical vision-language alignment to facilitate comprehensive coverage of gradually finer-grained surgical tasks, from visual perception, temporal analysis, to high-level reasoning.
Building upon this comprehensive dataset, we propose SurgVLM, which is built upon Qwen2.5-VL, and undergoes instruction tuning to 10+ surgical tasks. We further construct a surgical multimodal benchmark, SurgVLM-Bench, for method evaluation. SurgVLM-Bench consists of 6 popular and widely-used datasets in surgical domain, covering several crucial downstream tasks. Based on SurgVLM-Bench, we evaluate the performance of our SurgVLM (3 SurgVLM variants: SurgVLM-7B, SurgVLM-32B, and SurgVLM-72B), and conduct comprehensive comparisons with 14 mainstream commercial VLMs (e.g., GPT-4o, Gemini 2.0 Flash, Qwen2.5-Max).
Extensive experimental results show that the proposed SurgVLM consistently surpasses mainstream commercial VLMs.
SurgVLM-72B achieves 75.4\% improvement on overall arena score compared with Gemini 2.0 Flash, including a 96.5\% improvement in phase recognition, 87.7\% in action recognition, 608.1\% in triplet prediction, 198.5\% in instrument localization, 28.9\% in critical view safety detection, and 59.4\% in a comprehensive multi-task VQA dataset.
Beyond raw performance gains, SurgVLM demonstrates robust generalization and open‑vocabulary QA, establishing a scalable, accurate, and clinically reliable paradigm for unified surgical intelligence. Model weights, dataset, benchmark, and code will be available at \url{https://jinlab-imvr.github.io/SurgVLM}.

\end{abstract}}]

\let\thefootnote\relax\footnote{$\ast$ Corresponding Author ~~ $\dagger$ Equal Contribution}

\section{Introduction}
\begin{figure*}[htbp]  
	\centering
	\includegraphics[width=1\linewidth]{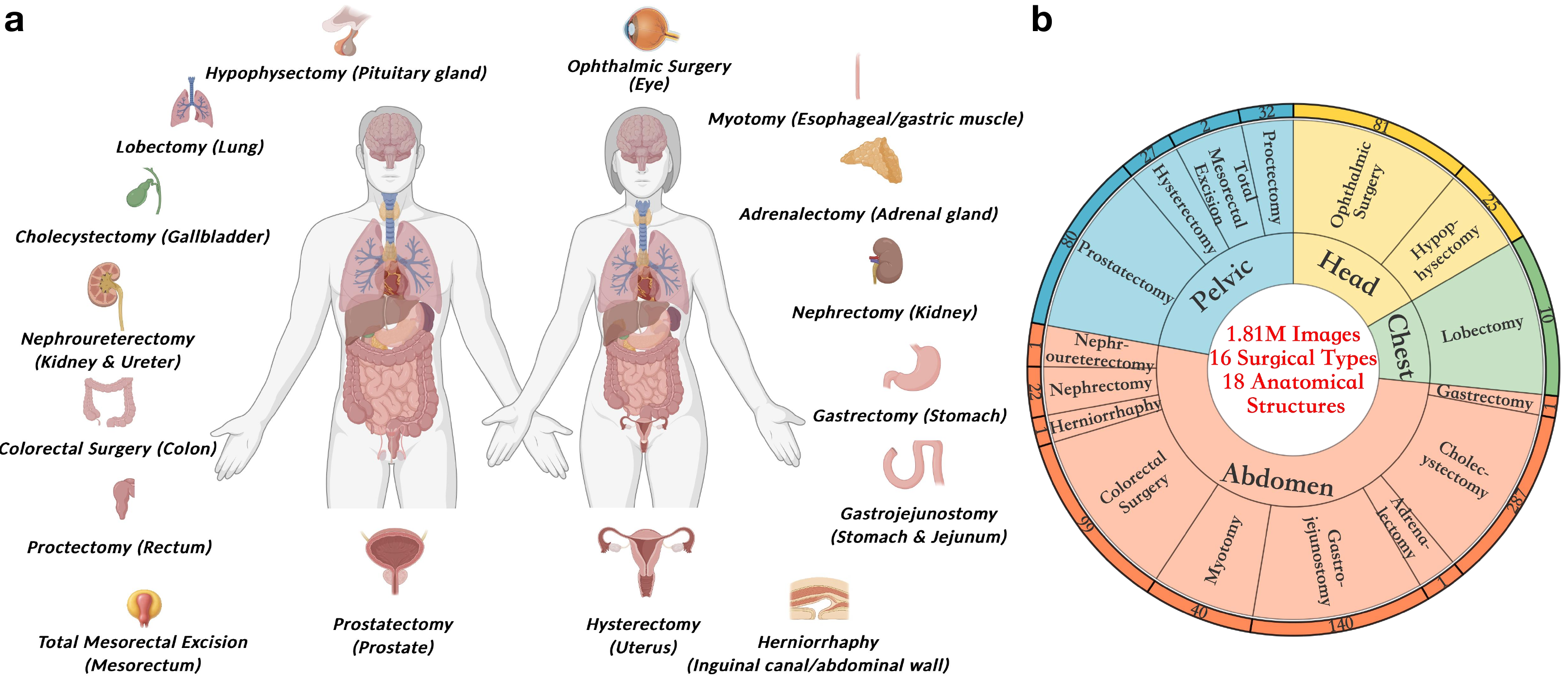}
    \caption{\textbf{Illustration of our Surgical Multimodal database SurgVLM-DB}. \textbf{a}, SurgVLM-DB contains 16 surgical types and 18 anatomical structures, reflecting the broad diversity of SurgVLM-DB.
    \textbf{b}, SurgVLM-DB contain 1.181M annotated images with 7.79M conversations, and the distribution of video numbers demonstrates large-scale comprehensive data in various surgical types, ensuring the robustness of SurgVLM.}
    \label{Figure:1}
\end{figure*}

\begin{figure*}
    \centering
    \includegraphics[width=0.9\linewidth]{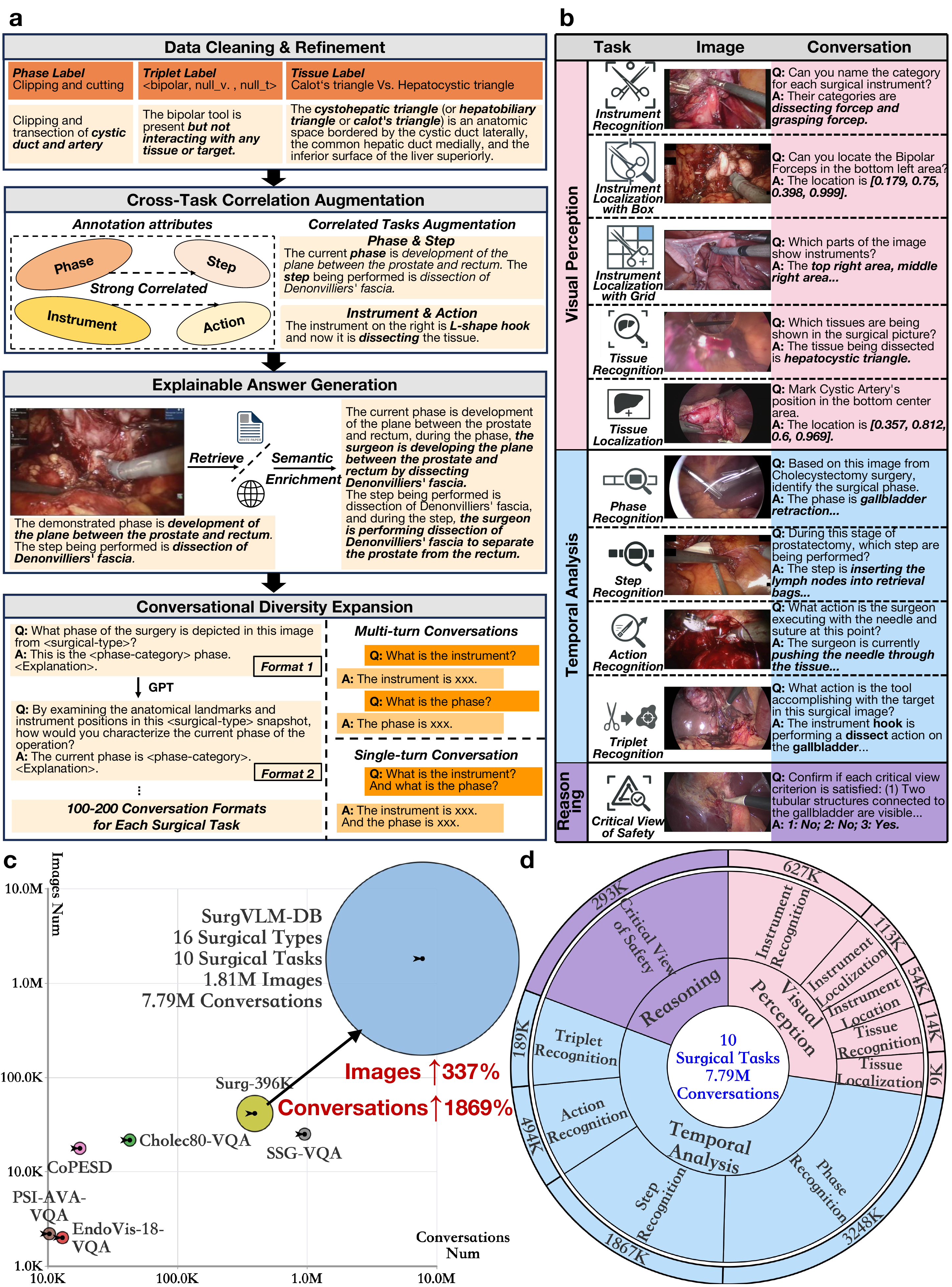}
    \caption{\textbf{Illustration of our Surgical Multimodal Database, SurgVLM-DB}.
    \textbf{a}, Construction pipeline can be divided into four modules.
    \textbf{b}, Overview of task hierarchy in SurgVLM-DB containing 10 surgical tasks, ranging from visual perception to temporal analysis to reasoning.
    \textbf{c}, Comparison showing SurgVLM-DB has the largest number of images, conversations, and surgical types indicated by the size of bubble to our knowledge.
    \textbf{d}, Distribution of conversations according to 10 surgical tasks.}
    \label{Figure:2}
\end{figure*}

\newcolumntype{Y}{>{\centering\arraybackslash}X}

\begin{table*}[htbp] \small
\centering
\caption{Comparison of our SurgVLM-DB with other surgical multimodal datasets. SurgVLP~\cite{yuan2023learning} is a collection of surgical lectures from Youtube without specific number of surgical types.}
\label{tab:datase comparison}
\begin{tabular}{c|ccccc}
\hline
\rowcolor{gray!30} Dataset      & Years & Surgical Types Num & Images          & Annotation Types & Conversations Num             \\ \hline


Cholec80-VQA~\cite{seenivasan2022surgical} & 2022  & 1           & 21.6K  & QA Pairs        & 43K          \\

EndoVis-18-VQA~\cite{seenivasan2022surgical} & 2022  & 1           & 2K  & QA Pairs        & 11.8K          \\

EndoVis-VQLA~\cite{bai2023surgical} & 2023  & 1           & 2.2K   & QA Pairs; Bbox  &9.5K           \\
PSI-AVA-VQA~\cite{seenivasan2023surgicalgpt}  & 2024  & 1           & 2.2K   & QA Pairs   &     10.3K           \\
SurgVLP*~\cite{yuan2023learning} & 2023 & N\textgreater1 & 25.5K & QA Pairs & 25.5K\\
CoPESD~\cite{wang2024copesd}       & 2024  & 1           & 17.7K   & QA Pairs; Bbox   & 17.7K         \\
SSG-VQA~\cite{yuan2024advancing} & 2024  &    1        & 25K  & QA Pairs        & 960K          \\
Surg-396K \cite{wang2025endochat}    & 2025  & 3   & 41.4K   & QA Pairs; Bbox       & 396K      \\

\rowcolor{green!30} SurgVLM-DB (Ours)    & 2025  & 16   & 1810K   & QA Pairs; Bbox       & 7798.4K      \\\hline
\end{tabular}
\end{table*}

Foundation models pre-trained on large-scale datasets have demonstrated remarkable success across various domains, particularly in healthcare. For example, in radiology, these models have enhanced tasks such as chest X-ray diagnosis~\cite{liu2023radiologyllama2, wu2023towards, zheng2024large}, automated generation of radiology reports~\cite{zhang2023knowledge, paschali2025foundation}, and identification and localization of abnormalities~\cite{liu2024expert}. In pathology, these models have advanced applications, including histopathological image analysis~\cite{chen2024towards, wang2024pathology}, tumor grading and staging~\cite{lu2024visual, vorontsov2024foundation}, and biomarker prediction~\cite{lu2024multimodal, ding2024multimodal, vaidya2025molecular}. These models extend beyond traditional medical image analysis to encompass text-based understanding and multimodal integration. By leveraging diverse modalities such as medical imaging, electronic health records (EHRs), publications, and clinical notes, foundation models provide a holistic understanding of medical contexts. Their impact has been transformative, improving clinical workflows, enhancing and streamlining patient care.

Despite these advancements, the application of foundation models in the surgical domain remains largely underexplored. 
Surgical interventions involve complex workflows with highly intricate and dynamic surgical scenes characterized by i) limited contrast and ambiguous boundaries among anatomical structures; ii) unstable lighting conditions and visual obstructions from smoke and blood; and iii) reduced visibility due to a limited field of view, frequent tool and camera motions, and tool occlusions.
Intelligent analysis of complex surgical procedures is essential to provide cognitive assistance to surgeons, encompassing diverse yet interconnected vision tasks, such as instrument and tissue recognition, surgical phase and action recognition, and critical view of safety assessment. These tasks form the foundational capabilities of modern Computer-Assisted Intervention (CAI) systems, facilitating context-aware decision-making, and enhancing coordination and communication among surgical team members. Moreover, beyond intraoperative use, these can significantly advance surgical educational training, improve surgical report documentation and postoperative patient monitoring.

Developing a surgical foundation model to address these versatile tasks can systematically and comprehensively support the entire surgical workflow. By seamlessly integrating multimodal data, including visual scenes and textual instructions, such a vision-language foundation model can effectively model task correlations. It can also leverage contextual information, including prior interactions and dialogue logs of other relevant tasks, to provide precise cognitive assistance.
For example, the foundation model first possesses the fundamental \textit{visual perception} capability, for instrument recognition, instrument localization, tissue recognition, and tissue localization~\cite{twinanda2016endonet,Nwoye2022Rendezvous,wang2024copesd,mascagni2024endoscapes2023,he2024pitvqa,seenivasan2022surgical,psychogyios2024sarrarp50,wang2022autolaparo, ghamsarian2024cataract,luengo20222020cataract,alabi2024cholecinstanceseg,carstens2023dresden,allan20192017,jimenez2022surgical,ayobi2023pixel,hu2024ophnet,liu2024surgical,nwoye2025cholectrack20}. 
Based on extracted visual semantics, the foundation model can achieve more complex tasks requiring \textit{temporal analysis}, such as surgical phase recognition, action recognition, and triplet recognition~\cite{bouget2015detecting, twinanda2016endonet,ayobi2023pixel,he2024pitvqa,seenivasan2022surgical,lavanchy2024multibypass140,hu2024ophnet,low2025surgraw}. 
Afterwards, the surgical foundation model can take these previous cues as context for high-level and fine-grained \textit{reasoning}, such as Critical View of Safety assessment~\cite{mascagni2024endoscapes2023} and error detection \cite{shao2024think}.
In addition to leveraging the synergy among these associated tasks, incorporating textual instructions enables the foundation model to understand surgeons' natural language commands, providing immersive and personalized assistance, ultimately aiming to achieve an enhanced patient safety with improved operative quality.

A key challenge in developing surgical foundation models is the absence of large-scale, high-quality, multimodal surgical datasets. General-purpose Vision-Language Models (VLMs)~\cite{bai2025qwen2.5vl, liu2023llava, hurst2024gpt, team2023gemini, wu2024deepseek, li2025baichuan, wang2024qwen2, ye2024mplug, zhang2024internlm, yao2024minicpm, abouelenin2025phi, zeng2023cognition, zhang2024long, team2025kimi}, trained predominantly on natural images and text, often exhibit inefficiency by generating excessive, clinically irrelevant outputs. Additionally, their outputs tend to be ambiguous, presenting multiple plausible scenarios rather than definitive, medically meaningful answers. Such ambiguity and verbosity undermine their alignment with surgeons' professional standards and real-world clinical requirements, significantly limiting their reliability and applicability in surgical practice. To address these limitations, it is highly desired to develop the specialized surgical foundation model capable of providing precise, hierarchical cognitive support, facilitating surgical decision-making, navigation, and task automation, thereby advancing toward the next generation of surgical operations.

To address this challenge, we construct SurgVLM-DB, a large-scale multimodal surgical database comprising over 1.81 million images with 7.79 million conversations, including 16 surgical types, 18 anatomical structures, and 10 crucial surgical tasks, from 23 publicly available datasets. 
To enhance generalization and mitigate biases, we develop an effective database construction pipeline, including Data Cleaning \& Refinement, Cross-Task Correlation Enrichment, Explainable Answer Generation, and Conversational Diversity Expansion. 
Notably, during correlation enrichment, we form hierarchically structured question-answer pairs to leverage the inherent relationships among different surgical tasks (fundamental visual perception with 5 tasks (instrument recognition, instrument localization with bounding boxes, instrument location with grid position, tissue recognition, tissue localization), to temporal analysis with 4 tasks (phase recognition, step recognition, action recognition, and triplet recognition), to high-level reasoning tasks (CVS assessment)). 
This structured database serves as the building block for the following learning strategy design of hierarchical vision-language alignment, enabling the model to learn not only from isolated task-specific annotations, but also from context-rich, interrelated surgical knowledge, improving both structured reasoning and response coherence. 


Building upon the constructed SurgVLM-DB, we propose SurgVLM, a family of large vision-language foundation models, designed to unify surgical AI tasks within a single, scalable framework. SurgVLM contain SurgVLM-7B, SurgVLM-32B, and SurgVLM-72B, which are model agnostic and can be built on other general-purpose VLMs. 
We propose a scalable training strategy that includes multimodal projector pretraining to hierarchically align surgical visual and textual representations, and mixed instruction tuning on surgical data to capture domain knowledge. This approach enables the model to implicitly learn task interdependencies and improves performance across interrelated vision–language tasks. 
Furthermore, the architecture of SurgVLM contains the processor of dynamic resolution for diverse sizes of surgical scenes and visual tokens compressing for effective LLM inference.

To evaluate SurgVLM's diverse abilities of visual perception, temporal analysis, and reasoning, we further construct a surgical multimodal benchmark, SurgVLM-Bench, for method evaluation. SurgVLM-Bench consists of 6 popular and widely-used datasets in surgical domain, covering several crucial downstream tasks. 
We evaluate SurgVLM on SurgVLM-Bench, and establish a benchmark of large VLMs on the surgical domain, including our proposed SurgVLM and 14 mainstream commercial VLM methods. 
Experimental results show that our SurgVLM-7B, SurgVLM-32B, and SurgVLM-72B achieve 32.3\%–198.5\% relative gains over these mainstream commercial VLM methods.
Comprehensive evaluations demonstrate that SurgVLM achieves superior performance over mainstream commercial VLMs. SurgVLM paves the way for safer, more efficient, and scalable AI-assisted surgery, setting a new standard for multimodal surgical intelligence.

Our main contributions can be summarized as:

1. We construct SurgVLM-DB, a large-scale multimodal surgical database comprising over 1.81 million images with 7.79 million conversations, including 16 surgical types, 18 anatomical structures, and 10 crucial surgical tasks.

2. We propose SurgVLM, a family of surgical multimodal foundation models designed to unify surgical tasks within a single, scalable framework, including SurgVLM-7B, SurgVLM-32B, and SurgVLM-72B.

3. We establish SurgVLM-Bench, a systematic multimodal evaluation benchmark with 6 popular surgical datasets, ranging from fundamental visual perception, temporal analysis, to high-level reasoning tasks.
\section{Results}
\subsection{SurgVLM-DB: Multimodal Surgical Database}
Developing robust generalist AI models for surgical applications requires large-scale, high-quality multimodal datasets encompassing diverse surgical scenarios. Unlike natural image datasets, the collection and annotation of a high-quality surgical database are significantly more challenging due to privacy concerns, the need for expert annotations, and the complexity of surgical workflows. 
Though the great contribution of existing public surgical databases to the community, they are still limited in scale, modality, or task diversity, making it difficult to develop the generalist VLMs. To address this gap, we construct SurgVLM-DB, a large-scale multimodal surgical database that integrates data from public surgical dataset repositories, covering a wide range of surgical tasks. 
As shown in Figure \ref{Figure:1}a, SurgVLM-DB contains 16 surgical types and 18 anatomical structures, reflecting the wide variability of the database.
Overall, SurgVLM-DB contains 1.181M annotated images with 7.79M conversations, covering 849 long-term surgical videos, and the distribution demonstrates large-scale comprehensive data in diverse surgical types, ensuring the robustness of SurgVLM (see Figure \ref{Figure:1}b). 
Furthermore, to effectively build the SurgVLM's diverse abilities of visual perception, temporal analysis, we design the hierarchical structure of the surgical tasks based on the domain knowledge. As shown in Figure \ref{Figure:2}b, an overview of the task hierarchy in SurgVLM-DB, ranging from visual perception to temporal analysis and reasoning. Tasks are organized based on increasing contextual dependency, task complexity, and clinical assistance.
The distribution of 7.79 million conversations in SurgVLM-DB according to 10 surgical tasks is shown in Figure \ref{Figure:2}d.
The comparison with existing datasets is illustrated in \ref{Figure:2}c, with the size indicated by the size of the bubble. To our best knowledge, SurgVLM-DB is an advanced surgical multimodal dataset with the largest number of images and conversations. More detailed comparison is presented in Table \ref{tab:datase comparison}.

\paragraph{\textbf{\textit{Data Construction Pipeline}}}
To systematically address the key challenges in constructing large‐scale surgical datasets—namely label inconsistency, fragmented task definitions, weak vision–language alignment, and overfitting on fixed prompt styles—we design a four‐stage pipeline (Figure \ref{Figure:2}a) in which each module builds on the output of the previous one:

\medskip
\noindent\textbf{Module 1: Data Cleaning \& Refinement.}  
The surgical labels are sometimes coarse (e.g.\ labeled as “other”). In addition, labels across different datasets are institution‐specific shorthand, or formatted inconsistently, leading to model confusion and tokenization misalignment. The ambiguous and inconsistent label is harmful for stable training of VLMs, causing misunderstanding for specific key words.
Hence, we conduct the \emph{Data Cleaning and Refinement} to replace each ambiguous label with an explicit, context‐rich description mapped to professional terms, and keeps semantic consistency by enforcing NIH‐standard~\cite{NIHwebsite} medical terminology. For example of removing ambiguous labels, we replace the \emph{“other”} with \emph{“The frame does not contain the defined phases, steps, or instruments."}. For example of keeping semantic consistency, we replace \emph{“Calot‘s triangle”} and \emph{“Hepatocystic triangle”} with \emph{“Cystohepatic Triangle”}. These clear and consistent surgical labels are beneficial for effective training of VLMs.

\vspace{-2mm}
\medskip
\noindent\textbf{Module 2: Cross-Task Correlation Enrichment.}  
Treating surgical tasks in isolation (e.g.\ phase vs.\ step vs.\ instrument) prevent models from learning their intrinsic dependencies. In this regard, we perform \emph{Cross-Task Correlation Enrichment}, pairing the labels of correlated attributes within a surgical scene into a single, holistic textual prompt, so the model can explicitly leverage hierarchical relationships among tasks. For example, given the strong correlation between phase and step, we combine them in one prompt as “In the current frame, the phase is \emph{developing the Space of Retzius} and the step is \emph{prevesical dissection}.” Similarly, recognizing the correlation between instrument and action, we generate prompts such as “The instrument on the right is \emph{L-shape hook} and now it is \emph{dissecting the tissue}.” 
These correlated prompts transform single-attribute annotations into joint queries, encouraging the model to infer one task from another rather than memorizing isolated labels.  
\medskip

\vspace{-2mm}
\medskip
\noindent\textbf{Module 3: Explainable Answer Generation.}  
The prompt annotation can be further enriched with semantics to provide a more detailed depiction of the surgical scene.
Therefore, we conduct \emph{Explainable Answer Generation} to augment each sample with more detailed explanations that tie visual evidence into the textual descriptions. 
As shown in Figure \ref{Figure:2}a, the annotation from last module is enriched to  
“The current phase is development of the plane between prostate and rectum, \emph{during which the surgeons separate the posterior prostatic fascia}. The current step is dissection of Denonvilliers’ fascia,\emph{ during which a clear surgical plane between prostate and rectum is created}.”  
This semantic enrichment can benefit fine‐grained vision-language alignment of local visual details by using such extended text description, enhancing the stability of VLMs in various surgical scenes. Furthermore, the explainable answer improves the trustworthiness of SurgVLM in real-world applications.

\vspace{-2mm}
\medskip
\noindent\textbf{Module 4: Conversational Diversity Expansion.}
Next, based on the rich and explainable description, we convert it to the conversation format by inserting the information into several templates. 
Note that Linguistic variety is of importance in VLM training, as the fixed question–answer templates can cause rapid overfitting in VLMs with large parameters.  
Therefore, we exploit a two-pronged strategy to maximize dialogue variability while preserving logical consistency, with typical examples shown in \ref{Figure:2}a. i) \emph{Conversation Format Variation.} We construct 100–200 distinct conversation templates by varying prompt phrasing, word ordering, and valid information density.
ii) \emph{Single‐turn and Multi‐turn Dialogues.} To further enrich interaction styles, each sample is presented in both Multi‐turn Dialogue and Single‐turn Dialogue. This aims to simulate the highly diverse conversation styles in actual human interaction with LLM systems. 
By interleaving varied prompt styles and turn structures, we maintain logical consistency while preventing memorization of any single format.

\medskip
\noindent\textbf{Inter‐module Flow.} Outputs from Module 1 (refined surgical labels) serve as inputs to Module 2 (correlated surgical tasks), whose paired instructions are then semantically expanded in Module 3, and finally diversified into robust multi‐turn dialogues in Module 4. Through multi-stage enhancement, our SurgVLM-DB more accurately reflects real clinical interactions by supporting diverse query formats, and offering clinically explainable assistance with enriched textual context by retrieving surgical knowledge and abstracting pixel-level visual semantics. More importantly, this strategy aligns heterogeneous procedures and tasks at the definition level, reducing model confusion and easing the burden of learning multiple tasks within a single model. At the same time, it preserves format diversity, preventing the model from overfitting to any specific surgery type or task.

\subsection{SurgVLM-Bench: Comprehensive Multimodal Evaluation Benchmark}

For model evaluation, we construct SurgVLM-Bench, a suite of six well-known surgical datasets spanning three levels of tasks, i.e., surgical visual perception, temporal analysis, and reasoning. 
With various tasks from basic instrument localization~\cite{allan20192017}, visual question answering (with instrument and tissue recognition, instrument localization as the major ones)~\cite{seenivasan2022surgical}, to phase recognition~\cite{twinanda2016endonet}, action recognition~\cite{psychogyios2024sarrarp50}, triplet recognition~\cite{Nwoye2022Rendezvous}, and CVS assessment~\cite{mascagni2024endoscapes2023}, our SurgVLM-Bench can comprehensively validate the performance of each method. 
More details of the datasets in SurgVLM-Bench are shown in the supplementary.

Based on it, we evaluate our proposed SurgVLM and compare it with 14 mainstream commercial VLMs. 
All datasets in SurgVLM-Bench employ the official training/testing splits and are balanced in terms of surgical types and anatomical structures, ensuring a comprehensive, end‑to‑end evaluation of multimodal surgical intelligence without fine-tuning.
\vspace{-15pt}
\paragraph{\textit{Overall Setup and Evaluation Metrics.}}
By default, all comparative baseline models and our SurgVLM‑72B variant are evaluated under a multiple‑choice question (MCQ) protocol, where the model selects the correct answer from a predefined set of options. To better reflect the ability of SurgVLM in more realistic surgical scenarios of visual perception, temporal analysis, and reasoning, we further evaluate SurgVLM‑7B, SurgVLM‑32B, and SurgVLM‑72B under an Open‑Vocabulary (OV) protocol. In this setting, models generate format-free responses instead of selecting from limited choices, allowing a more faithful assessment of flexible, context-aware reasoning, an essential capability for clinical deployment. 
Moreover, 
in the OV setting, responses must contain the exact ground truth keyword. Paraphrases or synonyms of these keywords are marked incorrect and this makes the evaluation challenging. This strict requirement on precise keywords is critical in our surgical domain, so that the system can provide professional and accurate assistance to surgeons and enhance patient safety.

Regarding the evaluation metrics, we first employ the arena score, a commonly-used overall evaluation metric, for comprehensive validation in the leaderboard.
This composite metric normalizes each dataset’s raw performance (6 benchmarks in total) to a 0–100 scale and then sums them to yield a single score out of 600. By aggregating performance across all datasets, the arena score provides a unified, overall evaluation for holistic comparison.

Moreover, we also employ the official evaluation metrics from the corresponding datasets for detailed and direct comparison.
All benchmarks employ the official training/testing splits and are balanced in terms of surgical types and anatomical structures. 
More details of the evaluation metrics for SurgVLM-Bench are shown in supplementary.

\vspace{-15pt}
\paragraph{\textit{Benchmarking in Visual Perception Tasks}} We first evaluate the visual perception level tasks, including instrument localization, instrument recognition and tissue recognition. 
EndoVis2017 dataset~\cite{allan20192017} is employed for instrument localization, which contains ten porcine abdominal procedures recorded with da Vinci Xi systems and annotated with bounding boxes for ten common laparoscopic instruments. Models are mainly scored by mean Intersection-over-Union (mIoU) across the held-out test split, quantifying their precision in grounding each instrument within complex surgical scenes. 
In addition, we use EndoVis2018-VQA~\cite{seenivasan2022surgical} for the evaluation of instrument and tissue recognition as well as instrument localization. Endovis2018-VQA comprises over 3,000 image–question pairs spanning instrument, tissue and procedural queries. Performance is reported as answer accuracy against the ground truth.

\vspace{-15pt}
\paragraph{\textit{Benchmarking in Temporal Analysis Tasks}} 
Three datasets are utilized in SurgVLM-Bench to assess model’s ability to interpret surgical workflows. 
We use Cholec80 dataset~\cite{twinanda2016endonet}  for surgical phase recognition, which contains 80 laparoscopic cholecystectomy videos. Models are required to select the correct phase over seven predefined surgical phases and are mainly evaluated by frame‐level accuracy. 
We use SAR-RARP dataset~\cite{psychogyios2024sarrarp50}  for action recognition, which annotates eight fine‐grained robotic actions related to suturing and cutting during prostatectomy. Model performance is mainly measured by classification accuracy. 
We further use CholecT50 dataset~\cite{Nwoye2022Rendezvous} for triplet recognition, where each frame is annotated with instrument–verb–target triplets (covering 11 instruments, 15 verbs and 12 targets); we compute mean average precision as the main metric for evaluation.

\vspace{-15pt}
\paragraph{\textit{Benchmarking in Reasoning Tasks}} 
We further benchmark on the high-level reasoning task, i.e., critical view of safety (CVS) assessment, by using the Endoscape2023 dataset~\cite{mascagni2024endoscapes2023}. Endoscape2023 samples 255 frames from laparoscopic cholecystectomy videos and labels each frame for three CVS criteria, (1) exposure of the cystic plate, (2) clearance of the lower third of the gallbladder, and (3) identification of only two structures entering the gallbladder. Models predict each criterion as a binary label, and we report overall accuracy on classifying whether the frame is at CVS or not by considering all three criteria. This evaluation measures a model’s ability to verify surgical safety protocols rather than simply recognize visual patterns.

\subsection{Performance Comparison of our SurgVLM with Other Mainstream VLMs}

\begin{figure*}[htbp]  
	\centering
	\includegraphics[width=0.92\linewidth]{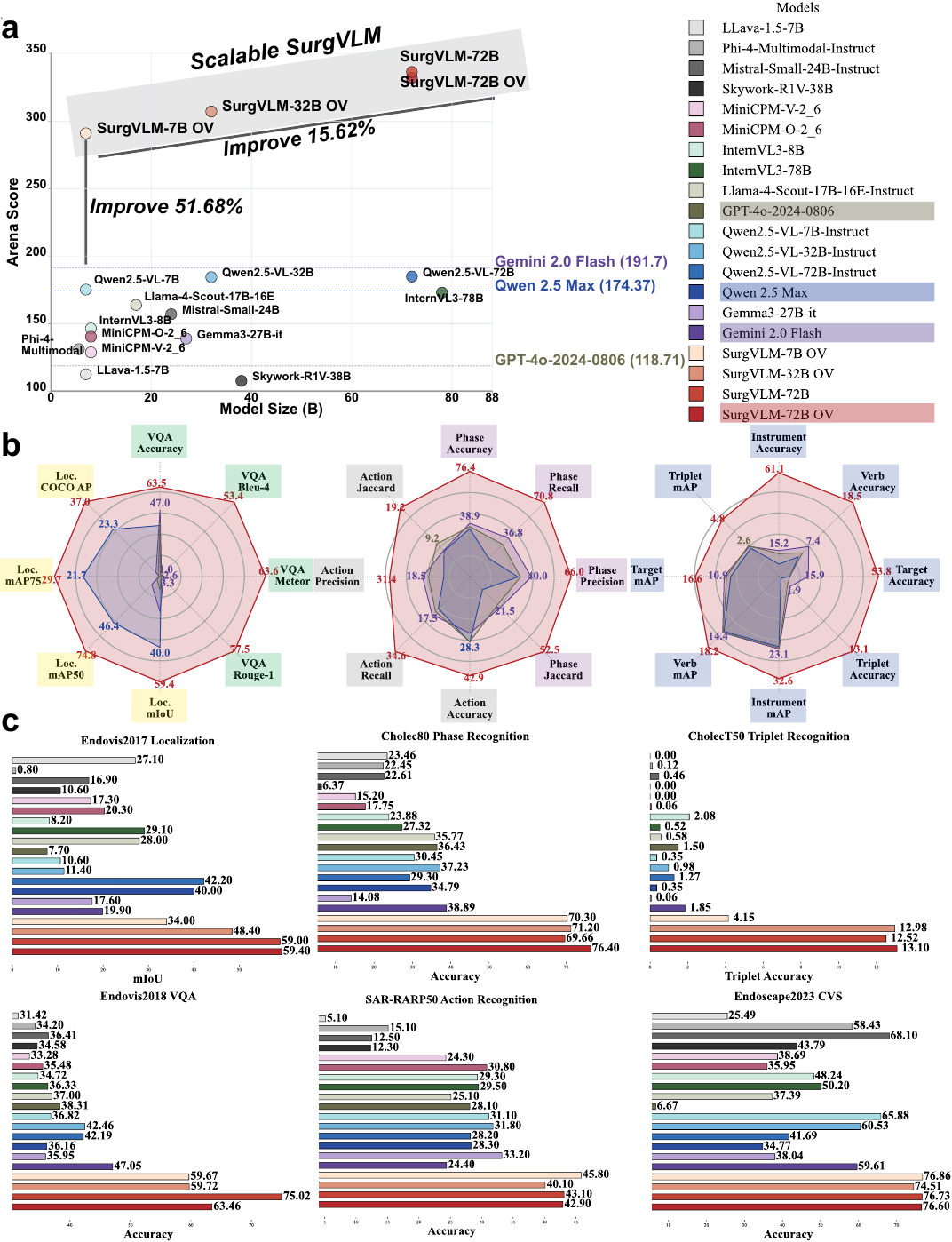}
	\caption{\textbf{Comparison of SurgVLM and 14 mainstream commercial VLMs}. \textbf{a}, Leaderboard on SurgVLM-Bench by overall arena score, demonstrating superior performance of our SurgVLM compared with other mainstream commercial VLMs.
    \textbf{b}, Comprehensive comparison with Gemini 2.0 Flash, Qwen2.5 Max, and GPT-4o. SurgVLM consistently outperforms commercial VLMs across 24 metrics.
    \textbf{c}, Detailed comparison with 14 mainstream commercial VLMs on the most important metrics of six popular surgical datasets. SurgVLM models achieve state-of-the-art performance across all metrics}
    \label{Figure:Results}
\end{figure*}

\begin{table*}[!htbp]
\footnotesize
\centering
\setlength{\tabcolsep}{2pt}
\caption{Leaderboard of Vision-Language Models on SurgVLM-Bench.}
\label{tab:arena_extended_updated}
\begin{tabular}{lccccc}
\hline
\rowcolor{gray!30}
\textbf{Model} & \textbf{Institute} & \textbf{LLM} & \textbf{Vision Encoder} & \textbf{Evaluation} & \textbf{Arena Score $\uparrow$} \\
\hline
\rowcolor{green!30}
SurgVLM-72B Lora-tuning (Ours) 
    & iMVR Lab
    & Qwen2.5-72B~\cite{yang2024qwen2.5} 
    & DFN CLIP~\cite{fang2023data} 
    & MCQ & \cellcolor{green!30}336.21 \\
\rowcolor{green!20}
SurgVLM-72B Lora-tuning (Ours) 
    & iMVR Lab
    & Qwen2.5-72B~\cite{yang2024qwen2.5} 
    & DFN CLIP~\cite{fang2023data} 
    & OV & \cellcolor{green!20}331.86 \\
\rowcolor{green!10}
SurgVLM-32B Freeze-tuning (Ours) 
    & iMVR Lab
    & Qwen2.5-32B~\cite{yang2024qwen2.5} 
    & DFN CLIP~\cite{fang2023data} 
    & OV & \cellcolor{green!10}306.91 \\
\rowcolor{green!5}
SurgVLM-7B Full-tuning (Ours) 
    & iMVR Lab
    & Qwen2.5-7B~\cite{yang2024qwen2.5} 
    & SIGLIP2~\cite{tschannen2025siglip} 
    & OV & \cellcolor{green!5}290.78 \\
\rowcolor{red!10}
Gemini 2.0 Flash~\cite{team2023gemini} 
    & Google DeepMind
    & – 
    & – 
    & MCQ & 191.70 \\
Qwen2.5-VL-72B-Instruct~\cite{bai2025qwen2.5vl} 
    & Alibaba Cloud
    & Qwen2.5-72B~\cite{yang2024qwen2.5} 
    & DFN CLIP~\cite{fang2023data} 
    & MCQ & 184.85 \\
Qwen2.5-VL-32B-Instruct~\cite{bai2025qwen2.5vl} 
    & Alibaba Cloud
    & Qwen2.5-32B~\cite{yang2024qwen2.5} 
    & DFN CLIP~\cite{fang2023data} 
    & MCQ & 184.40 \\
Qwen2.5-VL-7B-Instruct~\cite{bai2025qwen2.5vl}  
    & Alibaba Cloud
    & Qwen2.5-7B~\cite{yang2024qwen2.5} 
    & DFN CLIP~\cite{fang2023data} 
    & MCQ & 175.20 \\
Qwen 2.5 Max~\cite{bai2025qwen2.5vl} 
    & Alibaba Cloud
    & – 
    & – 
    & MCQ & 174.37 \\
InternVL3-78B~\cite{zhu2025internvl3}  
    & Shanghai AI Lab
    & InternViT-6B-448px-V2\_5~\cite{chen2024expanding} 
    & Qwen2.5-72B~\cite{yang2024qwen2.5} 
    & MCQ & 172.97 \\
Llama-4-Scout-17B-16E-Instruct
    & Meta AI
    & – 
    & – 
    & MCQ & 163.84 \\
Mistral-Small-3.1-24B-Instruct
    & Mistral AI
    & – 
    & – 
    & MCQ & 156.98 \\
InternVL3-8B~\cite{zhu2025internvl3}  
    & Shanghai AI Lab
    & InternViT-300M-448px-V2\_5~\cite{chen2024expanding} 
    & Qwen2.5-7B~\cite{yang2024qwen2.5}  
    & MCQ & 146.42 \\
MiniCPM-O-2\_6~\cite{team2025minicpm-minicpm-v} 
    & ModelBest
    & – 
    & – 
    & MCQ & 140.34 \\
Gemma3-27B-it~\cite{team2025gemma} 
    & Google DeepMind
    & – 
    & SIGLIP~\cite{zhai2023sigmoid} 
    & MCQ & 138.93 \\
Phi-4-Multimodal-Instruct~\cite{abouelenin2025phi} 
    & Microsoft
    & Phi-4-Mini~\cite{abouelenin2025phi} 
    & SIGLIP~\cite{zhai2023sigmoid} 
    & MCQ & 131.10 \\
MiniCPM-V-2\_6~\cite{team2025minicpm-minicpm-v} 
    & MiniCPM Team
    & – 
    & – 
    & MCQ & 128.77 \\
GPT-4o-2024-0806~\cite{hurst2024gpt} 
    & OpenAI
    & – 
    & – 
    & MCQ & 118.71 \\
LLava-1.5-7B~\cite{liu2023llava} 
    & WAIV Lab
    & LLama2~\cite{touvron2023llama2} 
    & CLIP~\cite{radford2021learning} 
    & MCQ & 112.57 \\
Skywork-R1V-38B~\cite{peng2025skywork} 
    & Skywork AI
    & DeepSeek-R1-Distill-Qwen-32B~\cite{guo2025deepseek} 
    & – 
    & MCQ & 107.64 \\
\hline
\end{tabular}
\end{table*}

\begin{figure*}[htbp]  
	\centering
	\includegraphics[width=\linewidth]{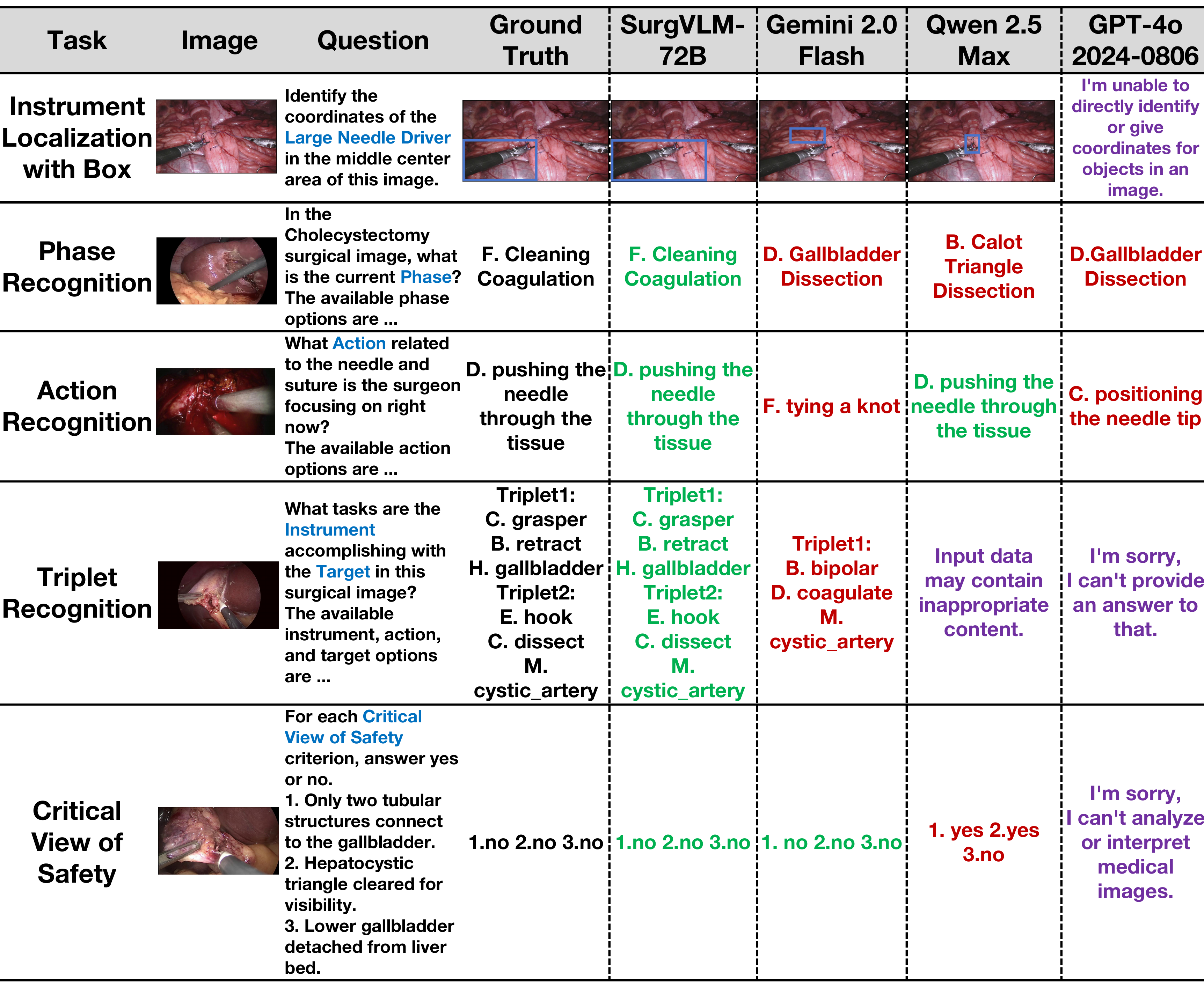}
	\caption{\textbf{Qualitative comparison with typical examples, of SurgVLM-72B, Gemini 2.0 Flash, Qwen 2.5 Max, and GPT-4o}, demonstrating the superior performance of SurgVLM. Green texts indicate correct response, while red texts represent wrong response. Security checking with purple color is one of the essential factors to cause unstable performance of these commercial VLMs.}
    \label{Figure:4}
\end{figure*}

\begin{table*}[!htbp] \small
\centering
\setlength{\tabcolsep}{3pt} 
\renewcommand{\arraystretch}{1.1} 
\caption{Performance on Cholec80 Phase Recognition, RARP Action Recognition, and CholecT50 Triplet Recognition.}
\label{tab:benchmark1}
\begin{tabularx}{\textwidth}{l Y *{8}{Y}}
\toprule
& & \multicolumn{4}{c}{Cholec80 Phase Recognition} & \multicolumn{4}{c}{RARP Action Recognition} \\
\cmidrule(lr){3-6}\cmidrule(lr){7-10}
Model & Eval & Accuracy & Recall & Precision & Jaccard & Accuracy & Recall & Precision & Jaccard \\
\midrule
LLava-1.5-7B~\cite{liu2023llava}                  & MCQ     & 23.46 & 14.22 & 11.53 & 6.31  & 5.10  & 12.50 & 0.64  & 0.64  \\
Phi-4-Multimodal-Instruct~\cite{abouelenin2025phi} & MCQ     & 22.45 & 14.83 & 12.16 & 5.55  & 15.10 & 12.83 & 11.41 & 3.47  \\
Mistral-Small-3.1-24B-Instruct-2503               & MCQ     & 22.61 & 19.87 & 26.55 & 9.92  & 12.50 & 12.68 & 5.89  & 1.78  \\
InternVL3-8B~\cite{zhu2025internvl3}    & MCQ     & 23.88 & 15.21 & 15.19 & 6.67  & 29.30 & 12.78 & 9.96  & 6.13  \\
InternVL3-78B~\cite{zhu2025internvl3}   & MCQ     & 27.32 & 24.03 & 30.48 & 13.30 & 29.50 & 12.87 & 22.82 & 6.28  \\
MiniCPM-V-2\_6~\cite{team2025minicpm-minicpm-v}    & MCQ     & 15.20 & 12.84 & 13.79 & 5.82  & 24.30 & 12.07 & 13.43 & 7.25  \\
MiniCPM-O-2\_6~\cite{team2025minicpm-minicpm-v}    & MCQ     & 17.75 & 19.68 & 29.24 & 9.22  & 30.80 & 13.13 & 20.12 & 5.74  \\
Gemma3-27B-it~\cite{team2025gemma}                & MCQ     & 14.08 & 17.49 & 19.67 & 4.45  & 33.20 & 13.86 & 18.01 & 5.32  \\
Skywork-R1V-38B~\cite{peng2025skywork}            & MCQ     & 6.37  & 14.67 & 13.32 & 1.38  & 12.30 & 13.02 & 19.83 & 3.08  \\
Llama-4-Scout-17B-16E-Instruct                    & MCQ     & 35.77 & 14.38 & 11.34 & 5.33  & 25.10 & 12.50 & 3.14  & 3.14  \\
Qwen2.5-VL-7B-Instruct~\cite{bai2025qwen2.5vl}   & MCQ     & 30.45 & 16.69 & 26.17 & 7.73  & 31.10 & 12.50 & 3.89  & 3.89  \\
Qwen2.5-VL-32B-Instruct~\cite{bai2025qwen2.5vl}  & MCQ     & 37.23 & 23.05 & 29.63 & 13.30 & 31.80 & 13.65 & 41.35 & 5.21  \\
Qwen2.5-VL-72B-Instruct~\cite{bai2025qwen2.5vl}  & MCQ     & 29.30 & 19.94 & 27.75 & 10.50 & 28.20 & 13.04 & 22.30 & 5.94  \\
GPT-4o-2024-0806~\cite{hurst2024gpt}                       & MCQ     & 36.43 & 31.02 & 32.99 & 19.55 & 28.10 & 15.99 & 16.15 & 9.16  \\
Qwen 2.5 Max~\cite{bai2025qwen2.5vl}              & MCQ     & 34.79 & 17.83 & 31.41 & 8.91  & 28.30 & 14.60 & 10.20 & 7.14  \\
\rowcolor{red!10} Gemini 2.0 Flash~\cite{team2023gemini}             & MCQ     & 38.89 & 36.76 & 40.03 & 21.47 & 24.40 & 17.51 & 18.54 & 7.27  \\
\midrule
\rowcolor{green!5}  SurgVLM-7B Full-tuning (Ours)       & OV & \textbf{70.30} & \textbf{61.90} & \textbf{59.76} & \textbf{43.86} & \textbf{45.80} & \textbf{31.32} & \textbf{35.07} & \textbf{20.68} \\
\rowcolor{green!10} SurgVLM-32B Freeze-tuning (Ours)    & OV & \textbf{71.20} & \textbf{67.87} & \textbf{63.61} & \textbf{48.89} & \textbf{40.10} & \textbf{30.46} & \textbf{34.64} & \textbf{18.09} \\
\rowcolor{green!20} SurgVLM-72B Lora-tuning (Ours)      & MCQ     & \textbf{69.66} & \textbf{68.10} & \textbf{68.70} & \textbf{50.18} & \textbf{43.10} & \textbf{25.21} & \textbf{24.61} & \textbf{15.16} \\
\rowcolor{green!30} SurgVLM-72B Lora-tuning (Ours)      & OV & \textbf{76.40} & \textbf{70.84} & \textbf{66.02} & \textbf{52.45} & \textbf{42.90} & \textbf{34.64} & \textbf{31.45} & \textbf{19.22} \\
\midrule
& & \multicolumn{8}{c}{CholecT50 Triplet Recognition} \\
\cmidrule(lr){3-10}
Model & Eval & \makecell{Inst.\\Acc} & \makecell{Verb\\Acc} & \makecell{Target\\Acc} & \makecell{Triplet\\Acc} & \makecell{Inst.\\mAP} & \makecell{Verb\\mAP} & \makecell{Target\\mAP} & \makecell{Triplet\\mAP} \\
\midrule
LLava-1.5-7B~\cite{liu2023llava}                   & MCQ     & 3.69  & 9.41  & 0.23  & 0.00 & 22.48 & 14.12 & 9.41  & 2.35 \\
Phi-4-Multimodal-Instruct~\cite{abouelenin2025phi} & MCQ     & 14.83 & 12.98 & 3.46  & 0.12 & 22.32 & 13.99 & 9.39  & 2.35 \\
Mistral-Small-3.1-24B-Instruct-2503                 & MCQ     & 9.81  & 8.02  & 5.48  & 0.46 & 22.43 & 14.05 & 9.41  & 2.35 \\
InternVL3-8B~\cite{zhu2025internvl3}    & MCQ     & 51.24 & 8.94  & 26.20 & 2.08 & 22.35 & 14.01 & 9.70  & 2.39 \\
InternVL3-78B~\cite{zhu2025internvl3}   & MCQ     & 38.14 & 9.52  & 3.29  & 0.52 & 22.74 & 14.08 & 9.44  & 2.36 \\
MiniCPM-V-2\_6~\cite{team2025minicpm-minicpm-v}     & MCQ     & 18.70 & 13.04 & 1.62  & 0.00 & 22.54 & 14.20 & 9.42  & 2.35 \\
MiniCPM-O-2\_6~\cite{team2025minicpm-minicpm-v}     & MCQ     & 24.99 & 7.79  & 5.71  & 0.06 & 22.49 & 14.02 & 9.40  & 2.35 \\
Gemma3-27B-it~\cite{team2025gemma}                  & MCQ     & 8.89  & 7.39  & 1.56  & 0.06 & 22.44 & 14.01 & 9.44  & 2.35 \\
Skywork-R1V-38B~\cite{peng2025skywork}              & MCQ     & 19.85 & 9.69  & 1.67  & 0.00 & 22.42 & 14.02 & 9.41  & 2.35 \\
Llama-4-Scout-17B-16E-Instruct                     & MCQ     & 9.46  & 6.17  & 9.64  & 0.58 & 22.73 & 14.22 & 9.64  & 2.37 \\
Qwen2.5-VL-7B-Instruct~\cite{bai2025qwen2.5vl}     & MCQ     & 10.21 & 4.90  & 5.54  & 0.35 & 22.67 & 14.14 & 9.47  & 2.36 \\
Qwen2.5-VL-32B-Instruct~\cite{bai2025qwen2.5vl}    & MCQ     & 27.81 & 7.91  & 3.12  & 0.98 & 22.79 & 14.68 & 10.09 & 2.53 \\
Qwen2.5-VL-72B-Instruct~\cite{bai2025qwen2.5vl}    & MCQ     & 32.66 & 7.91  & 5.25  & 1.27 & 23.38 & 14.45 & 10.86 & 2.71 \\
GPT-4o-2024-0806~\cite{hurst2024gpt}                         & MCQ     & 13.33 & 5.89  & 5.94  & 1.50 & 22.80 & 14.34 & 10.42 & 2.60 \\
Qwen 2.5 Max~\cite{bai2025qwen2.5vl}               & MCQ     & 7.21  & 4.85  & 5.94  & 0.35 & 22.40 & 14.12 & 9.59  & 2.39 \\
\rowcolor{red!10} Gemini 2.0 Flash~\cite{team2023gemini}              & MCQ     & 15.18 & 7.44  & 15.87 & 1.85 & 23.14 & 14.45 & 10.91 & 2.54 \\
\midrule
\rowcolor{green!5}  SurgVLM-7B Full-tuning (Ours)       & OV & \textbf{46.28} & \textbf{12.29} & \textbf{37.10} & \textbf{4.15}  & \textbf{22.44} & \textbf{14.04} & \textbf{9.31}  & \textbf{2.35} \\
\rowcolor{green!10} SurgVLM-32B Freeze-tuning (Ours)    & OV & \textbf{60.18} & \textbf{17.77} & \textbf{54.41} & \textbf{12.98} & \textbf{33.62} & \textbf{18.51} & \textbf{15.90} & \textbf{4.63} \\
\rowcolor{green!20} SurgVLM-72B Lora-tuning (Ours)      & MCQ     & \textbf{62.20} & \textbf{18.64} & \textbf{57.07} & \textbf{12.52} & \textbf{31.44} & \textbf{19.16} & \textbf{21.01} & \textbf{6.35} \\
\rowcolor{green!30} SurgVLM-72B Lora-tuning (Ours)      & OV & \textbf{61.11} & \textbf{18.47} & \textbf{53.78} & \textbf{13.10} & \textbf{32.59} & \textbf{18.17} & \textbf{16.64} & \textbf{4.83} \\
\bottomrule
\end{tabularx}
\end{table*}

\begin{table*}[!htbp] \small
\centering
\setlength{\tabcolsep}{2.4pt} 
\renewcommand{\arraystretch}{1.1} 
\caption{Performance of Various VLMs on Endovis2017 Localization, Endovis2018 VQA, and Endoscape 2023 CVS.}
\label{tab:benchmark2}
\begin{tabularx}{\textwidth}{l*{9}{Y}}
\toprule
\multirow{2}{*}{Model} & \multirow{2}{*}{Eval} & \multicolumn{4}{c}{Endovis2017 Localization} & \multicolumn{4}{c}{Endovis2018 VQA} \\
\cmidrule(lr){3-6} \cmidrule(lr){7-10}
&& mIoU & mAP@50 & mAP@75 & COCOAP & Accuracy & BLEU‑4 & METEOR & ROUGE‑1 \\
\midrule
LLava-1.5-7B~\cite{liu2023llava}                & MCQ        & 27.10 & 21.40 &  1.70 &  6.40 & 31.42 &  1.97 &  7.58 & 11.21 \\
Phi-4-Multimodal-Instruct~\cite{abouelenin2025phi} & MCQ        &  0.80 &  0.30 &  0.10 &  0.10 & 34.20 &  0.26 &  1.12 &  1.40 \\
Mistral-Small-3.1-24B-Instruct-2503            & MCQ        & 16.90 &  5.40 &  0.30 &  1.30 & 36.41 &  0.11 &  0.40 &  0.53 \\
InternVL3-8B~\cite{zhu2025internvl3}           & MCQ        &  8.20 &  3.50 &  0.00 &  0.70 & 34.72 &  1.03 &  4.56 &  4.54 \\
InternVL3-78B~\cite{zhu2025internvl3}          & MCQ        & 29.10 & 17.80 &  1.60 &  5.10 & 36.33 &  0.18 &  1.33 &  1.03 \\
MiniCPM-V-2\_6~\cite{team2025minicpm-minicpm-v} & MCQ        & 17.30 &  5.40 &  0.10 &  1.30 & 33.28 &  1.25 &  5.51 &  6.91 \\
MiniCPM-O-2\_6~\cite{team2025minicpm-minicpm-v} & MCQ        & 20.30 & 10.00 &  0.40 &  2.80 & 35.48 &  0.21 &  0.64 &  1.03 \\
Gemma3-27B-it~\cite{team2025gemma}             & MCQ        & 17.60 &  7.10 &  0.20 &  1.50 & 35.95 &  0.03 &  0.12 &  0.27 \\
Skywork-R1V-38B~\cite{peng2025skywork}         & MCQ        & 10.60 &  1.40 &  0.00 &  0.30 & 34.58 &  0.31 &  1.20 &  1.47 \\
Llama-4-Scout-17B-16E-Instruct                 & MCQ        & 28.00 & 15.60 &  1.00 &  4.30 & 37.00 &  0.07 &  0.32 &  0.44 \\
Qwen2.5-VL-7B-Instruct~\cite{bai2025qwen2.5vl}  & MCQ        & 10.60 &  0.60 &  0.00 &  0.20 & 36.82 &  4.72 & 15.79 & 13.80 \\
Qwen2.5-VL-32B-Instruct~\cite{bai2025qwen2.5vl} & MCQ        & 11.40 &  1.90 &  1.00 &  1.00 & 42.46 &  0.20 &  2.13 &  1.73 \\
Qwen2.5-VL-72B-Instruct~\cite{bai2025qwen2.5vl} & MCQ        & 42.20 & 45.20 & 25.10 & 25.50 & 42.19 &  0.24 &  1.62 &  2.37 \\
GPT-4o-2024-0806~\cite{hurst2024gpt}           & MCQ        &  7.70 &  3.10 &  0.30 &  0.90 & 38.31 &  0.25 &  1.29 &  1.13 \\
Qwen 2.5 Max~\cite{bai2025qwen2.5vl}            & MCQ        & 40.00 & 46.40 & 21.70 & 23.30 & 36.16 &  0.28 &  0.93 &  1.21 \\
\rowcolor{red!10} Gemini 2.0 Flash~\cite{team2023gemini}           & MCQ        & 19.90 &  8.30 &  0.30 &  2.00 & 47.05 &  1.02 &  2.60 &  3.31 \\
\midrule
\rowcolor{green!5} SurgVLM‑7B Full-tuning (Ours)                    & OV    & 34.00 & 25.60 &  6.40 &  9.70 & \textbf{59.67} & \textbf{53.56} & \textbf{63.33} & \textbf{76.83} \\
\rowcolor{green!10} SurgVLM‑32B Freeze-tuning (Ours)                & OV    & 48.40 & 51.80 & 24.30 & 27.00 & \textbf{59.72} & \textbf{53.64} & \textbf{63.36} & \textbf{76.93} \\
\rowcolor{green!20} SurgVLM‑72B Lora-tuning (Ours)                  & MCQ        & 59.00 & 73.60 & 27.50 & 35.70 & \textbf{75.02} & \textbf{39.40} & \textbf{52.19} & \textbf{71.04} \\
\rowcolor{green!30} SurgVLM‑72B Lora-tuning (Ours)                  & OV    & 59.40 & 74.80 & 29.70 & 37.00 & \textbf{63.46} & \textbf{53.39} & \textbf{63.61} & \textbf{77.50} \\
\midrule
\multirow{3}{*}{Model} & \multirow{3}{*}{Eval} & \multicolumn{8}{c}{Endoscape 2023 CVS} \\
\cmidrule(lr){3-10}
 && \multicolumn{4}{c}{Accuracy} & \multicolumn{4}{c}{Balanced Accuracy} \\
\cmidrule(lr){3-6} \cmidrule(lr){7-10}
      &            & Average & Criterion 1 & Criterion 2 & Criterion 3 & Average & Criterion 1 & Criterion 2 & Criterion 3 \\
\midrule
LLava-1.5-7B~\cite{liu2023llava}               & MCQ        & 25.49 & 26.27 & 20.78 & 29.41 & 49.85 & 51.04 & 49.29 & 49.22 \\
Phi-4-Multimodal-Instruct~\cite{abouelenin2025phi} & MCQ        & 58.43 & 54.51 & 64.71 & 56.08 & 44.82 & 44.20 & 46.27 & 44.00 \\
Mistral-Small-3.1-24B-Instruct-2503           & MCQ        & 68.10 & 67.45 & 74.51 & 62.35 & 45.82 & 46.39 & 46.11 & 44.94 \\
InternVL3-8B~\cite{zhu2025internvl3} & MCQ & 48.24 & 40.00 & 53.33 & 51.37 & 50.92 & 54.29 & 51.59 & 46.89 \\
InternVL3-78B~\cite{zhu2025internvl3} & MCQ & 50.20 & 31.76 & 50.20 & 68.63 & 53.50 & 52.02 & 57.54 & 50.94 \\
MiniCPM-V-2\_6~\cite{team2025minicpm-minicpm-v}        & MCQ        & 38.69 & 32.16 & 38.04 & 45.88 & 42.42 & 45.35 & 39.68 & 42.22 \\
MiniCPM-O-2\_6~\cite{team2025minicpm-minicpm-v}        & MCQ        & 35.95 & 33.73 & 25.88 & 48.24 & 45.81 & 47.46 & 44.52 & 45.44 \\
Gemma3-27B-it~\cite{team2025gemma}                  & MCQ        & 38.04 & 24.71 & 18.82 & 70.59 & 50.24 & 50.00 & 50.71 & 50.00 \\
Skywork-R1V-38B~\cite{peng2025skywork}             & MCQ        & 43.79 & 27.45 & 50.20 & 53.73 & 49.13 & 50.22 & 50.56 & 46.61 \\
Llama-4-Scout-17B-16E-Instruct                & MCQ        & 37.39 & 24.31 & 17.25 & 70.59 & 49.37 & 49.21 & 48.89 & 50.00 \\
Qwen2.5-VL-7B-Instruct~\cite{bai2025qwen2.5vl}  & MCQ        & 65.88 & 56.08 & 82.35 & 59.22 & 49.56 & 48.97 & 50.00 & 49.72 \\
Qwen2.5-VL-32B-Instruct~\cite{bai2025qwen2.5vl} & MCQ        & 60.53 & 31.76 & 79.22 & 70.59 & 50.04 & 52.02 & 48.10 & 50.00 \\
Qwen2.5-VL-72B-Instruct~\cite{bai2025qwen2.5vl}& MCQ        & 41.69 & 25.10 & 29.80 & 70.20 & 47.22 & 50.26 & 41.67 & 49.72 \\
GPT-4o-2024-0806~\cite{hurst2024gpt}                      & MCQ        &  6.67 &  6.67 &  5.88 &  7.45 &  9.07 &  9.23 &  8.81 &  9.17 \\
Qwen 2.5 Max~\cite{bai2025qwen2.5vl}           & MCQ        & 34.77 & 21.96 & 17.25 & 65.10 & 45.03 & 44.44 & 44.52 & 46.11 \\
\rowcolor{red!10} Gemini 2.0 Flash~\cite{team2023gemini}         & MCQ        & 59.61 & 47.84 & 63.92 & 67.06 & 52.37 & 57.90 & 50.16 & 49.06 \\
\midrule
\rowcolor{green!5} SurgVLM‑7B Full-tuning (Ours)                   & OV    & \textbf{76.86} & \textbf{75.29} & \textbf{82.35} & \textbf{72.94} & \textbf{51.33} & \textbf{50.00} & \textbf{50.00} & \textbf{50.00} \\
\rowcolor{green!10} SurgVLM‑32B Freeze-tuning (Ours)                & OV    & \textbf{74.51} & \textbf{72.55} & \textbf{80.00} & \textbf{70.98} & \textbf{54.12} & \textbf{53.51} & \textbf{54.68} & \textbf{54.17} \\
\rowcolor{green!20}SurgVLM‑72B Lora-tuning (Ours)                          & MCQ        & \textbf{76.73} & \textbf{76.47} & \textbf{82.75} & \textbf{70.98} & \textbf{51.39} & \textbf{52.38} & \textbf{51.11} & \textbf{50.67} \\
\rowcolor{green!30} SurgVLM‑72B Lora-tuning (Ours)                      & OV    & \textbf{76.60} & \textbf{76.08} & \textbf{83.14} & \textbf{70.59} & \textbf{51.40} & \textbf{51.59} & \textbf{52.22} & \textbf{50.39} \\

\bottomrule
\end{tabularx}
\end{table*}

Our evaluation framework assesses three SurgVLM variants (7B Full-tuning, 32B Freeze-tuning and 72B Lora-tuning~\cite{hu2022lora}) across the six tasks of SurgVLM‑Bench. 
We benchmark these models against 16 commercial VLMs, such as Gemini 2.0 Flash~\cite{team2023gemini}, Qwen 2.5 Max~\cite{bai2025qwen2.5vl} and GPT‑4o~\cite{hurst2024gpt} as well as leading open-source VLMs under a unified protocol. 
We compare these methods with our proposed SurgVLM (SurgVLM-72B) using a multiple-choice question (MCQ) format that requires the model to select the correct answer from the given options. We further evaluate our proposed SurgVLM (all three variants) under the more challenging open-vocabulary VQA setting, which requires free-form, context-aware responses.

As shown in Figure \ref{Figure:Results}a, the leaderboard of VLMs on SurgVLM-Bench, demonstrates superior performance of SurgVLM compared with other mainstream commercial VLMs. 
The radar plots in Figure \ref{Figure:Results}b show the comprehensive comparison with Gemini 2.0 Flash, Qwen2.5 Max, and GPT-4o. SurgVLM consistently outperforms commercial VLMs across all 24 metrics. 
Detailed comparison with 14 mainstream commercial VLMs on the most important metrics across six surgical tasks is shown in Figure \ref{Figure:Results}c. All SurgVLM models achieve state-of-the-art performance. Additionally, the detailed numerical comparison of aggregate arena scores is reported in Table \ref{tab:arena_extended_updated}, and the quantitative results for each task’s evaluation metrics are presented in Tables \ref{tab:benchmark1} and \ref{tab:benchmark2}.

\vspace{-15pt}
\paragraph{\textit{Overall Arena Score Comparison on Leaderboard}} The aggregate arena scores are illustrated in Figure \ref{Figure:Results}a, with details presented in Table \ref{tab:arena_extended_updated}, summing each model’s performance across all six SurgVLM‑Bench tasks for proprietary and open source VLMs alongside our SurgVLM variants. Among proprietary models, Gemini 2.0 Flash leads with 191.70 followed by Qwen 2.5 Max at 174.37 and GPT‑4o at 118.71. 
Open source VLMs perform similarly or worse, with Qwen2.5‑VL‑Instruct variants ranging from 175.20 (7B) to 184.85 (72B) and InternVL3‑78B (172.97), Llama‑4‑Scout‑17B (163.84) and Mistral‑Small‑3.1‑24B (156.98) falling below these marks. 
Many open source VLMs exhibit a strong bias toward a single choice under MCQ evaluation, inflating their scores without reflecting true surgical scene understanding and proprietary APIs sometimes truncate or refuse outputs due to security filters, causing inconsistent behavior.

All SurgVLM models are Supervised Fine-tuning (SFT) exclusively on SurgVLM‑DB, achieving substantially higher arena scores of 290.78 for SurgVLM-7B (Full-tuning) under OV evaluation, 306.91 for SurgVLM-32B (Freeze-tuning) under OV evaluation, 331.86 for SurgVLM-72B (Lora-tuning) MCQ evaluation, and 336.21 for SurgVLM-72B (Lora-tuning) under OV evaluation, as shown in Table~\ref{tab:arena_extended_updated}. 
The performance of SurgVLM increases in lockstep with model capacity and tuning strategy. Moving from full-tuning to freeze-tuning to LoRA adaptation yields successive gains, and the SurgVLM-72B under MCQ evaluation delivers an additional 4.35 point boost over its MCQ counterpart by demonstrating true OV reasoning. These results confirm that large-scale high-quality multimodal surgical data combined with targeted fine tuning is essential to unlock the potential to achieve robust, clinically-reliable surgical intelligence.

\vspace{-15pt}
\paragraph{\textit{Comparison on Specific Tasks}} At the visual perception level, we evaluate the model’s ability to recognize and localize surgical instruments and tissues, which requires fine-grained visual grounding and accurate concept recognition.
On the EndoVis2017 instrument localization benchmark, measured by mIoU, performance improves markedly from 19.90\% for Gemini to 59.00\% (72B MCQ) and 59.40\% (72B OV) across our four SurgVLM variants, as presented in Table~\ref{tab:benchmark2}.
For the EndoVis2018-VQA benchmark for instrument and tissue recognition, instrument localization, our models again show substantial gains. Accuracy increases from 47.05\% for Gemini to 75.02\% (72B MCQ) and 63.4\% (72B OV) for the SurgVLM variants, as shown in Table~\ref{tab:benchmark2}. These results show that generic VLMs lack the domain‑specific vocabulary and spatial acuity for surgical scenes, whereas our SurgVLM models demonstrate consistent and clinically meaningful performance.
Additionally, we can also evaluate instrument and tissue recognition tasks on CholecT50 dataset. 
As shown in Table~\ref{tab:benchmark1}, Gemini 2.0 Flash scores 15.18\% in accuracy of the instrument recognition, whereas SurgVLM achieves 62.20\% (72B MCQ) and 61.11\% (72B OV).
For the target (tissue) recognition task, the performance of Gemini 2.0 in accuracy is 15.87\%, compared to SurgVLM’s 57.07\% (72B MCQ) and 53.78\% (72B OV).

At the temporal analysis level, we evaluate the performance on tasks that require interpreting procedural context. 
Cholec80 phase recognition accuracy improves from 38.89\% for Gemini to 69.66\% (72B MCQ) and 76.40\% (72B OV), as shown in Table~\ref{tab:benchmark1}. RARP action recognition also rises from 24.40\% to 43.10\% (72B MCQ) and 42.90\% (72B OV), as shown in Table~\ref{tab:benchmark1}. 
For triplet recognition on CholecT50, verb accuracy jumps from 7.44\% to 18.64\% (72B MCQ) and 18.47\% (72B OV), and triplet accuracy increases from 2.08\% to 12.52\% (72B MCQ) and 13.10\% (72B OV), as shown in Table~\ref{tab:benchmark1}. 

For tasks requiring reasoning-level ability, we validate the results on critical view of safety assessment using Endoscape2023 dataset. As shown in Table~\ref{tab:benchmark2}, the average accuracy of 59.61\% for Gemini 2.0 Flash is raised to 76.86\% (7B), 74.51\% (32B), 76.73\% (72B MCQ) and 76.60\% (72B OV). 
Successful CVS assessment requires holistic anatomical understanding and protocol adherence, and the large performance improvement of SurgVLM demonstrates its promising potential to support high-stakes clinical decision making.

When evaluated in an OV format, mainstream commercial VLMs often produce long generic sentences and lack professional and precise keywords in surgery, highlighting their limited ability to provide effective assistance. An example is shown below. 
\begin{itemize}
  \item Prompt: What is the phase in this image of prostatectomy?
  \item SurgVLM's response: The current phase is \textbf{development of the plane between the prostate and rectum}, during the phase, the surgeon is developing the plane between the prostate and rectum by dissecting Denonvilliers' fascia.
  \item GPT-4o's response: “Based on the provided intraoperative image from a robot-assisted radical prostatectomy, the scene corresponds to the \textbf{Neurovascular Bundle (NVB) Dissection \underline{or} Apical Dissection Phase}. This determination is supported by the following observations. 1. Active use of two surgical instruments in close proximity to vascular and glandular tissues. 2. Visible bleeding control and meticulous dissection, indicative of work near sensitive anatomical zones such as the NVB. 3. Presence of fatty and fibrous tissues consistent with the neurovascular bundle region. In standardized multi-phase classification systems for prostatectomy (e.g., Montsouris or Vattikuti), this frame most likely corresponds to: Phase 5–6 in the Vattikuti 9-step classification: Neurovascular Bundle Preservation or Apex Dissection. Further confirmation can be made if the specific phase taxonomy used in the dataset is provided.”
\end{itemize}
GPT-4o unclearly provides two possible answers with ambiguous analysis, while SurgVLM outputs clear and precise surgical phase with an explainable description. Hence, SurgVLM is reliable foundation model supporting open-vocabulary conversation towards real-world application.

\subsection{Scalability of our Surgical Foundation Model} 
As SurgVLM scales from 7 billion to 32 billion to 72 billion parameters, we gradually reduce the fraction of weights updated during training, drawing insights from successful scaling strategies demonstrated in mainstream commercial VLMs \cite{zhu2025internvl3, bai2025qwen2.5vl}. Specifically, the SurgVLM-7B model is fully fine-tuned, updating all parameters to effectively learn detailed surgical vision-language alignment. For the larger SurgVLM-32B model, only a small subset of parameters is fine-tuned, with most parameters frozen; this limited parameter updating strategy helps avoid overfitting caused by excessively adapting the large capacity to a relatively small surgical vision-language dataset. 
Finally, the largest SurgVLM-72B model utilizes a LoRA adapter-based fine-tuning strategy, further restricting parameter updates to compact adapter modules. This enables the model to leverage its extensive pre-trained capacity efficiently, maintaining strong generalization while achieving precise surgical vision-language alignment. 


\vspace{-15pt}
\paragraph{\textit{On fundamental visual perception tasks,}} parameter scaling yields significant and consistent improvements, particularly for simpler tasks such as localization, which rely heavily on basic visual capabilities. However, the effectiveness of scaling law is still different between surgical visual perception tasks

Instrument localization performance on EndoVis2017 dramatically improves from an mIoU of 34.00\% (7B) to 48.40\% (32B), and further advances notably to 59.40\% (72B OV), as illustrated in Table~\ref{tab:benchmark2}. This highlights that larger models, pretrained extensively on general visual perception data from natural scenes, greatly enhance the ability to accurately localize surgical instruments. In contrast, for tasks requiring fine-grained surgical vision-language alignment, such as instrument and tissue recognition, scaling the parameter count yields diminishing returns beyond a certain point. Specifically, recognition accuracy on CholecT50 initially sees substantial improvement from 46.25\% (7B) to 60.18\% (32B) for instruments, and from 37.10\% (7B) to 54.41\% (32B) for tissues. However, performance plateaus at larger scales, reaching 61.11\% (instruments) and slightly decreasing to 53.78\% (tissues) at 72B OV. 

These observations indicate a clear scaling law for fundamental visual localization tasks, where each increment in model size consistently improves performance due to inherent general visual perception abilities. Conversely, tasks reliant on detailed surgical vision-language alignment exhibit significant performance improvements with initial parameter scaling but encounter marginal gains beyond mid-sized models (around 30B parameters). This phenomenon suggests that enhancing performance on recognition tasks is more dependent on high-quality, fine-grained surgical domain-specific data rather than merely increasing model size. Practically, a mid-sized model around 30B parameters with targeted adapter tuning can thus achieve competitive localization accuracy with considerably lower computational costs compared to a fully fine-tuned 72B model.

\vspace{-15pt}
\paragraph{\textit{At the temporal analysis level,}} relatively simpler classification tasks such as phase and action recognition exhibit more modest improvements as the model scale increases. Specifically, on the Cholec80 benchmark, phase recognition accuracy progresses from 70.30\% (7B) to 71.20\% (32B), and further reaches 76.40\% (72B OV). However, performance notably decreases to 69.99\% with SurgVLM-72B MCQ, likely due to the constraints of the multiple-choice evaluation format. Similarly, for RARP action recognition, accuracy initially drops from 45.80\% (7B) to 40.10\% (32B), then partially recovers to 43.10\% (72B MCQ) and 42.90\% (72B OV).

In contrast, the more complex triplet recognition task on CholecT50 shows significantly greater gains from model scaling. Accuracy substantially increases from 4.15\% (7B) to 12.98\% (32B), stabilizes at 12.52\% (72B MCQ), and achieves 13.10\% (72B OV). This pronounced improvement reflects the underlying influence of enhanced visual perception capabilities, as the triplet composition involves \textit{Instrument}, \textit{Action}, and \textit{Tissue}. As indicated in Table~\ref{tab:benchmark1}, the enhancements from 7B to 32B models are predominantly driven by considerable improvements in \textit{Instrument Recognition} and \textit{Tissue Recognition} (Target Recognition). Notably, the gains in \textit{Action Recognition} (Verb Recognition) are relatively minor, supporting the observation that improvements in triplet recognition primarily arise from more accurate identification of instruments and tissues, rather than actions.



\vspace{-15pt}
\paragraph{\textit{At the reasoning level,}} the average accuracy of the CVS task remains consistently high across different model scales, achieving 76.86\% (7B), 74.51\% (32B), 76.73\% (72B MCQ), and 76.60\% (72B OV), respectively. It is important to emphasize that CVS represents a notably challenging reasoning task, as it requires the model to perform fine-grained vision-language alignment utilizing the SurgVLM-DB for accurate decision-making. However, due to the scarcity of available CVS data, even after applying our meticulously designed data pipeline to enhance data quantity and diversity, the resulting dataset remains insufficient to fully exploit the potential scaling law of larger models. 
Consequently, this plateau in performance indicates that once essential anatomical knowledge and workflow contexts are learned through fine-tuning, mid-sized models are already capable of effectively integrating the three binary safety checks into coherent judgments. Further improvements are more likely to emerge from enriching and diversifying the CVS dataset rather than simply increasing model size.

\subsection{Qualitative Comparison} 
We further illustrate the qualitative comparison with typical examples in Figure \ref{Figure:4}. We compare our SurgVLM-72B with Gemini 2.0 Flash, Qwen 2.5 Max, and GPT-4o, on tasks including instrument localization, phase recognition, action recognition, triplet recognition, and CVS assessment.  
Green texts indicate the correct response, while red texts represent the wrong response. 
Overall, our SurgVLM demonstrates the superior performance with more accurate and complete outputs.
Commercial VLMs sometimes output security checking shown in purple color, which is one of the essential factors leading to their unstable performance. 
For instrument localization, three mainstream commercial VLMs output completely incorrect bounding boxes or reject the response, while SurgVLM-72B outputs very accurate bounding boxes. 
For triplet recognition, it is difficult for other VLMs to identify complete triplets. For example, Gemini 2.0 Flash outputs single triplet, while our SurgVLM-72B can output both triplets correctly.

\section{Discussion}

To build a surgical intelligence in real world, we need to review the key points of building foundation model in the specific domain, including large-scale high-quality database, diverse tasks aligned with applications in the real world, and scalable models with dynamic training strategies. We have discovered several principles to build surgical VLMs as follows.

\vspace{-15pt}

\paragraph{\textit{Extensive Surgical Knowledge in SurgVLM-DB}}
Fine-tuning general vision-language models (VLMs) on SurgVLM-DB facilitates efficient and precise adaptation to surgical workflows. With carefully constructed and structured annotations, SurgVLM-DB enables the adaptation of mainstream VLMs to the surgical domain by employing dynamic supervised fine-tuning. Our models trained on SurgVLM-DB:

\begin{enumerate}
\item Generate more concise outputs with fewer tokens, improving the effectiveness of VLMs in surgical settings. This results in faster responses in the operating environment, enhancing the interaction between the surgeon and AI. Furthermore, the models can be deployed on portable surgical devices, reducing the computational resource requirements in hospitals (e.g., by utilizing less expensive CPUs or GPUs with lower memory capacity). Similar strategies for efficiency have been successfully demonstrated in mainstream commercial VLMs such as BLIP-2~\cite{li2023blip2} and Flamingo~\cite{alayrac2022flamingo}, where fine-tuning on specialized domains enhances both the speed and relevance of outputs in real-world applications.

\item Achieve higher performance across six diverse datasets in SurgVLM-Bench, underscoring the foundational principle of building a domain-specific vision-language foundation model. This principle is supported by previous studies in the natural scene domain, where task-specific fine-tuning allows large language models (LLMs) to focus on more concise knowledge representations, effectively bridging domain gaps in vision-language models \cite{zhu2025internvl3}. Such targeted fine-tuning of the LLM decoder leads to substantial improvements in surgical applications.

\item Integrate professional surgical terminology to ensure real-world clinical relevance. The adaptation process aims to mirror the progression of medical trainees. Pretrained LLMs retain broad linguistic capabilities, while targeted fine-tuning on SurgVLM-DB imparts the fine-grained vision-language expertise necessary for advanced surgical AI assistance. This approach is akin to the adaptation strategies used in medical VLMs like LLava-med~\cite{li2023llava} and MedGemma~\cite{medgemma2025}, where domain-specific fine-tuning enables effective deployment in specialized fields, enhancing model accuracy and interpretability.
\end{enumerate}

\vspace{-15pt}
\paragraph{\textit{General Surgical Visual Representation}} 
How to effectively leverage the extensive knowledge of existing LLMs and VLMs to develop the surgical foundation model is important. 
While advances in vision–language architectures have often focused on ever‐larger backbones, our SurgVLM-Bench results reveal that off-the-shelf general‐purpose encoders already furnish rich, surgery-relevant visual features. 
Models such as DFN CLIP~\cite{fang2023data}, without surgical‐domain self‐supervised pretraining, can already provide strong support of feature extraction for developing surgical foundation models, 
mirroring observations in natural scenes where CLIP‐style encoders generalize broadly to downstream tasks \cite{radford2021learning, li2023blip2}. 
This indicates that the principal adaptation challenge lies not in crafting novel visual backbones but more in supplying high‐quality semantic supervision: Our precise, structured textual annotations enable the detailed instruction tuning, which can better map surgical visual semantics into the LLM’s latent space, therefore unlocking the latent representational power of existing encoders, as shown in Flamingo’s gains from semantic alignment over architectural changes \cite{alayrac2022flamingo, jia2021scaling}.

\vspace{-15pt}
\paragraph{\textit{Handling Diversity and Synergy Across Surgical Types}} 
Another core component in developing surgical foundation model is how to deal with diverse surgical types.
First, certain similar surgical types, such as those involving the same anatomical region, tend to exhibit high correlation.
For example, they share visual commonalities of tissue textures and similar functional cues in surgical instruments. 
Building on insights from general‐purpose VLM, where multi‐dataset pretraining on overlapping object categories enhances feature transfer and downstream performance \cite{radford2021learning}, our SurgVLM-DB unifies data from multiple related surgical types. With the joint training regime, our model can effectively exploit inherent cross‐type synergies, and yield richer visual representations and improve accuracy on different tasks.
On the other hand, most surgical types and procedures are different and exhibit distinct characteristics. In real-world scenarios, surgeons are typically aware of the surgical type to be performed. 
To simulate this prior knowledge and also enable a universal foundation model to perform well across diverse procedures, we include this prior knowledge, i.e., category of surgical type, into the prompt in instruction tuning.

\vspace{-15pt}
\paragraph{\textit{Relationship of Hierarchical Surgical Tasks}}
In clinical surgery, decision‐making hinges on continuous, accurate interpretation of complex scenes, temporal patterns, and cognitive assessments. Accordingly, visual perception, temporal analysis, and reasoning in a unified framework offer substantial advantages over treating these tasks in isolation, including enhanced end‐to‐end optimization \cite{radford2021learning}, real‐time robustness \cite{alayrac2022flamingo}, and alignment with clinical workflows \cite{twinanda2016endonet}. 
Inspired by integrated perception‐to‐reasoning pipelines in the general domain where early visual accuracy drives downstream temporal and cognitive performance \cite{radford2021learning, alayrac2022flamingo}, we propose a hierarchical surgical AI paradigm in training data grounded in two principles: (1) precise instrument and tissue recognition can benefit phase and action recognition quality \cite{twinanda2016endonet}; and (2) robust temporal analysis is essential for dependable decision‐making. 
These interdependencies underscore the necessity of co‐training various tasks together. 
Furthermore, this hierarchy naturally supports a curriculum learning strategy \cite{bengio2009curriculum, platanios2020competence}, starting with visual perception (instrument/tissue recognition and localization), advancing to temporal analysis (phase/action recognition), and culminating in reasoning tasks (CVS assessment). This strategy can improve data efficiency \cite{bengio2009curriculum}, accelerate convergence \cite{platanios2020competence}, and ensure overall model robustness across surgical workflow.

Overall, we propose SurgVLM, one of the first large vision-language foundation models for surgical intelligence, where this single universal model can tackle versatile surgical tasks.
To enable this, we construct a large-scale multimodal surgical database, SurgVLM-DB, comprising over 1.81 million frames with 7.79 million conversations, spanning more than 16 surgical types and 18 anatomical structures. 
We unify and reorganize 23 public datasets across 10 surgical tasks, followed by standardizing labels and doing hierarchical vision-language alignment to facilitate comprehensive coverage of gradually finer-grained surgical tasks, from visual perception, temporal analysis, to high-level reasoning.
Building upon this comprehensive dataset, we propose SurgVLM, which is built upon Qwen2.5-VL, and undergoes instruction tuning to 10 surgical tasks. 
We further construct a surgical multimodal benchmark, SurgVLM-Bench, for method evaluation. SurgVLM-Bench consists of 6 popular and widely-used datasets in surgical domain, covering several crucial downstream tasks. 
Based on SurgVLM-Bench, we evaluate the performance of our SurgVLM (3 SurgVLM variants: SurgVLM-7B, SurgVLM-32B, and SurgVLM-72B), and conduct comprehensive comparisons with 14 mainstream commercial VLMs (e.g., GPT-4o, Gemini 2.0 Flash, Qwen2.5-Max).
Extensive experimental results show that the proposed SurgVLM consistently surpasses state-of-the-art mainstream commercial VLMs.

Despite these advances, several major challenges remain.
Firstly, the scale and diversity of the database need to be expanded to leverage the scaling laws of larger models and enhance robustness. Yet, high-quality surgical annotations are costly and require clinical expertise. To address this, we envision a cyclic bootstrapping pipeline: using zero-shot SurgVLM to generate pseudo-labels, refining them through human-in-the-loop correction, and iteratively improving model performance.
Another limitation is the absence of a dedicated temporal dependency modeling module. Capturing temporal dynamics is important for surgical tasks, however, how to effectively extract the temporal cues when developing large vision-language model remains challenging.
We found that simply concatenating visual tokens and using positional encodings for temporal relationships, though computationally efficient, often yields insufficient results. 
Future work involves designing a multimodal fusion–based temporal dependency computation module. This module will integrate features such as inter-frame spatio-temporal representations, depth variations, and optical flow to accurately model temporal dependencies. 

Looking ahead, SurgVLM models, SurgVLM-DB and SurgVLM-Bench could lay the groundwork for broader surgical AI systems, not only supporting surgical decision making and education with strong perception and reasoning capability, but also assisting surgical planning and subtask autonomy. We hope our hierarchical benchmark and foundation model will catalyze more research in multimodal surgical AI, ultimately advancing safer, smarter, more efficient and accessible surgical care for the augmented hospital in future.

{
    \small
    \bibliographystyle{ieeenat_fullname}
    \bibliography{bibliography}
}

\clearpage
\onecolumn
\appendix
\noindent\LARGE{\textbf{Supplement}}
\setcounter{table}{0}
\renewcommand{\thetable}{S\arabic{table}}
\setcounter{figure}{0}
\renewcommand{\thefigure}{S\arabic{figure}}
\setcounter{section}{0}
\normalsize

\begin{figure*}[!htbp]  
	\centering
	\includegraphics[draft=false, scale=0.55]{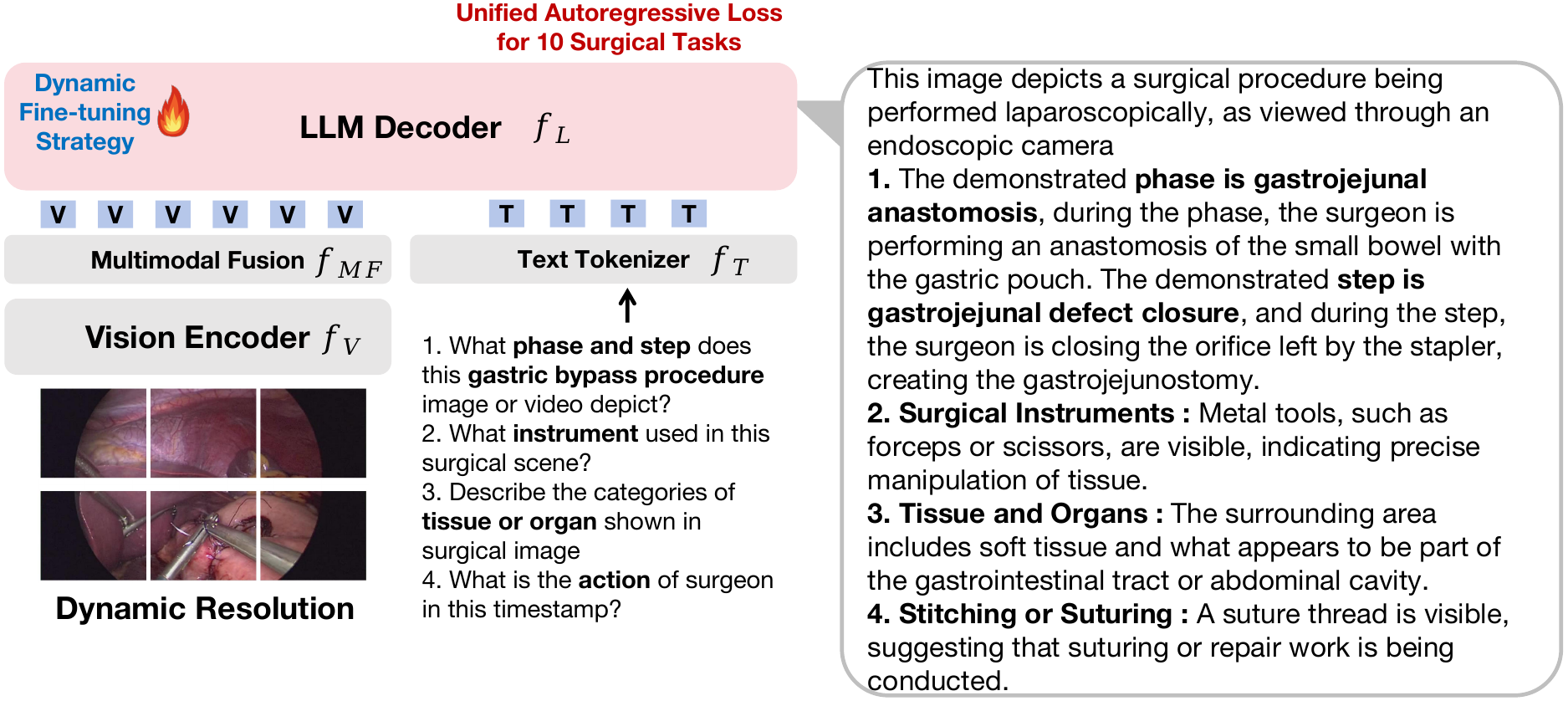}
	\caption{\textbf{Overall illustration of proposed SurgVLM.} SurgVLM model follows the Qwen2.5-VL architecture, consisting of a vision encoder, a multimodal token fusion module, and a large language model (LLM) decoder with unified autoregressive loss for 10 surgical tasks and dynamic fine-tuning strategy for effective training with different size of parameters.}
    \label{Figure:Model}
\end{figure*}

\section{Method}
In this section, we describe the underlying architecture of our model and the techniques employed to enhance its capabilities for surgical AI applications.

\subsection{Model Architecture}
Our \textbf{SurgVLM} model follows the Qwen2.5-VL architecture, consisting of a vision encoder, a multimodal token fusion module, and a large language model (LLM) decoder, as shown in Figure \ref{Figure:Model}. The vision encoder is a Transformer-based image backbone that processes images (and video frames) at their native resolution, while the LLM (initialized from Qwen2.5 LLM) serves as the decoder for generating outputs. We detail each component below.

\subsubsection{Vision Encoder with Dynamic Resolution}
The vision encoder is a re-designed Vision Transformer (ViT) that supports \emph{dynamic resolution} input. Instead of resizing images to a fixed size, the model maps an input image of arbitrary size to a variable-length sequence of visual tokens. Concretely, an input image
\[
I \in \mathbb{R}^{H\times W\times 3}
\]
is first divided into non‐overlapping patches of size $P\times P$ (with $P=14$ pixels in our design). Each patch is flattened and linearly projected into a $d_v$-dimensional embedding to form an initial token. If $P_p$ denotes the $p$-th patch (flattened to $\mathbb{R}^{P^2\times 3}$), then
\begin{equation}
\mathbf{z}_p = W_e\,\mathrm{vec}(P_p) + \mathbf{b}_e,
\end{equation}
where $W_e\in\mathbb{R}^{d_v\times (P^2\cdot 3)}$ and $\mathbf{b}_e\in\mathbb{R}^{d_v}$ are the patch embedding weights. The resulting sequence $\{\mathbf{z}_p\}_{p=1}^N$ (with $N = \frac{H}{P}\cdot\frac{W}{P}$ patches) is then fed into Transformer layers with multi‐head self‐attention. To keep computation feasible for high‐resolution images (large $N$), we use \emph{window‐based} attention: self‐attention is applied only within local windows of size $M\times M$ patches, reducing complexity from $O(N^2)$ to $O(N\cdot M^2)$. Periodically, full‐image (global) attention layers are inserted to allow cross‐window information flow. This mechanism, inspired by Swin Transformer’s shifted windows, preserves spatial locality while optimizing inference speed. Each Transformer block’s attention is computed as
\begin{equation}
\mathrm{Att}(Q,K,V) \;=\; \mathrm{Softmax}\!\Bigl(\tfrac{QK^\top}{\sqrt{d_k}}\Bigr)\,V,
\end{equation}
where $Q,K,V$ are the query, key, and value matrices from the previous layer’s hidden states. We use $h$ attention heads, each of dimension $d_k$, and employ FFNs with SwiGLU activation and RMSNorm normalization in every layer.

\paragraph{Multimodal Positional Encoding (M-RoPE).}
To handle arbitrary image and video sizes without fixed positional embeddings, we adopt \emph{Multimodal Rotary Position Embedding} (M-RoPE). We decompose each token’s position into a triplet $(t,u,v)$ for temporal index $t$ (frame index in video, or $t=0$ for an image) and spatial patch coordinates $(u,v)$. The $d_v$-dimensional token is split into three sub-vectors,
\[
\mathbf{q} = \bigl[\mathbf{q}^{(t)};\,\mathbf{q}^{(h)};\,\mathbf{q}^{(w)}\bigr]
\quad\text{and}\quad
\mathbf{k} = \bigl[\mathbf{k}^{(t)};\,\mathbf{k}^{(h)};\,\mathbf{k}^{(w)}\bigr].
\]
We apply RoPE separately to each sub-vector:
\begin{equation}
\begin{aligned}
\mathbf{q}^{(t)\prime} &= \mathrm{RoPE}\bigl(\mathbf{q}^{(t)},t\bigr), &
\mathbf{q}^{(h)\prime} &= \mathrm{RoPE}\bigl(\mathbf{q}^{(h)},u\bigr), &
\mathbf{q}^{(w)\prime} &= \mathrm{RoPE}\bigl(\mathbf{q}^{(w)},v\bigr),
\end{aligned}
\end{equation}
and then concatenate to form $\mathbf{q}'=[\mathbf{q}^{(t)\prime};\,\mathbf{q}^{(h)\prime};\,\mathbf{q}^{(w)\prime}]$, with an analogous process for $\mathbf{k}'$. Here $\mathrm{RoPE}(\mathbf{x},n)$ rotates each feature pair in $\mathbf{x}$ by an angle proportional to $n$. For text tokens, we set $t=u=v$ equal to the token index, reducing M-RoPE to standard 1D RoPE. Moreover, SurgVLM aligns the temporal index $t$ to actual frame timestamps (in seconds) rather than simple frame counts, enabling the model to capture real elapsed time between frames.

\subsubsection{Multimodal Token Fusion}
After the vision Transformer, an image (or video) yields $N$ visual tokens. Feeding all $N$ tokens into the LLM is costly; we therefore apply an MLP‐based compression, grouping every four spatially adjacent patches into one token. Let
\[
\{\mathbf{z}_{i1},\mathbf{z}_{i2},\mathbf{z}_{i3},\mathbf{z}_{i4}\}
\]
be the embeddings of the $i$-th 2×2 patch block. We form
\[
\mathbf{x}_i = [\mathbf{z}_{i1};\,\mathbf{z}_{i2};\,\mathbf{z}_{i3};\,\mathbf{z}_{i4}]\in\mathbb{R}^{4d_v},
\]
then compute
\begin{equation}
\mathbf{h}_i = W_2\,\sigma\bigl(W_1\,\mathbf{x}_i + \mathbf{b}_1\bigr) + \mathbf{b}_2,
\end{equation}
where $W_1\in\mathbb{R}^{m\times 4d_v}$, $W_2\in\mathbb{R}^{d_\ell\times m}$, and $\sigma(\cdot)$ is the activation function. This reduces token count by 4× and projects into the LLM’s $d_\ell$-dimensional space. The fused tokens $\{\mathbf{h}_i\}$ are prepended to the text sequence for the Transformer decoder to process jointly.

\subsection{Hierarchical Vision-Language Alignment with Multi-Task Objective}

We adapt the pretrained SurgVLM (Qwen2.5-VL) to ten distinct surgical tasks by formulating them as a unified sequence-to-sequence modeling problem, optimized via a single autoregressive language modeling loss. Central to this multi-task objective is achieving hierarchical vision-language alignment at two levels:

\paragraph{Inter-task Hierarchical Alignment} The ten surgical tasks are grouped into three primary hierarchical categories, enabling the model to capture and leverage task dependencies across different levels of abstraction:

\begin{enumerate}
\item \textbf{Visual Perception}: Instrument Recognition; Instrument Localization with Box; Instrument Localization with Grid; Tissue Recognition; Tissue Localization.
\item \textbf{Temporal Analysis}: Phase Recognition; Step Recognition; Action Recognition; Triplet Recognition.
\item \textbf{Reasoning}: Critical View of Safety Assessment.
\end{enumerate}

This hierarchical structuring ensures that the model learns correlations between tasks, effectively interpreting comprehensive multi-task scene descriptions to enhance single-task inference capabilities.

\paragraph{Intra-task Hierarchical Alignment} Within each primary category, further hierarchical vision-language alignment is achieved. The model learns fine-grained associations between visual cues and corresponding textual instructions, capturing nuanced relationships crucial for accurate single-task execution and cross-task reasoning.

Each training instance consists of a serialized "prompt + label" pair, where prompts represent visual input via compressed visual tokens, and labels correspond to the ground-truth token sequences specific to each task. Mixing all tasks within each minibatch fosters shared representation learning, promoting robust inter- and intra-task alignment.

We use a unified autoregressive language modeling loss over all ten tasks. Let $\mathcal{D}$ denote the combined dataset from all tasks. For an example $(x, y) \in \mathcal{D}$, where $x$ is the input token sequence and $y=(y\_1, \dots, y\_L)$ the target sequence, the model parameters $\Phi$ are optimized by minimizing:

\begin{equation}
  \mathcal{L}_{\text{MT}}(\Phi) = - \sum_{(x, y) \in \mathcal{D}} \sum_{i=1}^{L} \log P_{\Phi} \left( y_i \mid y_{<i}, x \right)
\end{equation}

This unified loss enables the model to align vision and language representations across tasks holistically, supporting both fine-grained interpretation and task-level reasoning.

\subsection{Effective Training Paradigms for Scalable VLM}

To efficiently adapt SurgVLM for hierarchical vision-language alignment across diverse surgical tasks, we investigate three parameter-efficient training paradigms:

\begin{itemize}
\item \textbf{Full Fine-Tuning}: Update all parameters $\Phi$ comprehensively, facilitating deep integration of visual perception, temporal analysis, and reasoning capabilities.
\item \textbf{Freeze-Tuning}: Freeze the majority of model parameters (e.g., LLM decoder), selectively training only vision-fusion adapters to refine hierarchical alignment efficiently.
\item \textbf{LoRA-Tuning}: Integrate and train low-rank adapter matrices within Transformer layers, enabling targeted parameter adjustments for scalable performance.
\end{itemize}

All paradigms are trained using the same unified multi-task loss $\mathcal{L}\_{\text{MT}}$, enabling fair comparisons of training efficiency and alignment performance across the ten surgical tasks.

\begin{table}[htbp]
\centering
\caption{Configuration of SurgVLM-7B, SurgVLM-32B, and SurgVLM-72B.}
\begin{tabular}{lccc}
\toprule
\textbf{Configuration} 
  & \textbf{SurgVLM-7B} 
  & \textbf{SurgVLM-32B} 
  & \textbf{SurgVLM-72B} \\ 
\midrule
\multicolumn{4}{c}{\textbf{Vision Transformer (ViT)}} \\ 
\midrule
Hidden Size                      & 1280 & 1280                  & 1280 \\
\# Layers                        & 32   & 32                    & 32 \\
\# Num Heads                     & 16   & 16                    & 16 \\
Intermediate Size                & 3456 & 3456                  & 3456 \\
Patch Size                       & 14   & 14                    & 14 \\
Window Size                      & 112  & 112                   & 112 \\
Full Attention Block Indexes     & \{7,15,23,31\} & \{7,15,23,31\} & \{7,15,23,31\} \\ 
\midrule
\multicolumn{4}{c}{\textbf{Vision-Language Merger}} \\  
\midrule
In Channel                       & 1280 & 1280                  & 1280 \\
Out Channel                      & 3584 & 5120                  & 8192 \\ 
\midrule
\multicolumn{4}{c}{\textbf{Large Language Model (LLM)}} \\ 
\midrule
Hidden Size                      & 3584   & 5120                 & 8192 \\
\# Layers                        & 28     & 64                   & 80 \\
\# KV Heads                      & 4      & 8                    & 8 \\
Head Size                        & 128    & 128                  & 128 \\
Intermediate Size                & 18944  & 27648                & 29568 \\
Embedding Tying                  & \ding{55} & \ding{55}        & \ding{55} \\
Vocabulary Size                  & 151646 & 152064               & 151646 \\
\bottomrule
\end{tabular}
\label{table:config}
\end{table}

\section{Database Composition}
\begin{table}[htbp]
  \centering
  \footnotesize
  \caption{Detailed composition of large-scale database SurgVLM-DB grouped by surgical types}
  \label{tab:datasets_by_types}
  \begin{tabular}{@{}l l@{}}
    \toprule
    \textbf{Types} & \textbf{Datasets} \\
    \midrule
    Hysterectomy
      & AutoLaparo \cite{wang2022autolaparo}; LRSP-VQA \\
    Ophthalmic Surgery
      & Cataract-1K \cite{ghamsarian2024cataract}; CaDISv2 \cite{luengo20222020cataract}; OphNet \cite{hu2024ophnet} \\
    Cholecystectomy
      & Cholec80 \cite{twinanda2016endonet}; Cholec80-CVS \cite{rios2023cholec80cvs}; Cholec80-VQA \cite{seenivasan2022surgical}; CholecInstanceSeg \cite{alabi2024cholecinstanceseg}; CholecT50 \cite{Nwoye2022Rendezvous}; Endoscapes2023 \cite{mascagni2024endoscapes2023} \\
    Submucotomy
      & CoPESD \cite{wang2024copesd} \\
    Myotomy
      & DSAD \cite{carstens2023dresden} \\
    Nephrectomy
      & EndoVis2017 \cite{allan20192017}; EndoVis2018 \cite{jimenez2022surgical}; EndoVis2018-VQA \cite{seenivasan2022surgical}; EndoVis2018-VQA-Triplet \cite{seenivasan2022surgical}; LRSP-VQA~\cite{du2024lsrp} \\
    Prostatectomy
      & GraSP \cite{ayobi2023pixel}; PSA-AVA-VQA \cite{seenivasan2022surgical}; SAR-RARP \cite{psychogyios2024sarrarp50}; SurgCoTBench \cite{low2025surgraw}; LRSP-VQA~\cite{du2024lsrp} \\
    Gastrojejunostomy
      & MultiBypass140 \cite{lavanchy2024multibypass140} \\
    Pituitary Surgery
      & PitVQA \cite{he2024pitvqa} \\
    Colorectal Surgery
      & SegCol \cite{ju2024segcol} \\
    Lobectomy
      & SurgCoTBench \cite{low2025surgraw}; LRSP-VQA~\cite{du2024lsrp} \\
    Total Mesorectal Excision
      & LRSP-VQA~\cite{du2024lsrp} \\
    Nephroureterectomy
      & LRSP-VQA~\cite{du2024lsrp} \\
    Gastrectomy
      & LRSP-VQA~\cite{du2024lsrp} \\
    Herniorrhaphy
      & LRSP-VQA~\cite{du2024lsrp} \\
    Adrenalectomy
      & LRSP-VQA~\cite{du2024lsrp} \\
    \bottomrule
  \end{tabular}
\end{table}

As shown in Table \ref{tab:datasets_by_types}, our SurgVLM-DB contain 16 surgical types with their corresponding publicly available datasets: Hysterectomy (AutoLaparo \cite{wang2022autolaparo}, LRSP-VQA \cite{du2024lsrp}), Ophthalmic Surgery (Cataract-1K \cite{ghamsarian2024cataract}, CaDISv2 \cite{luengo20222020cataract}, OphNet \cite{hu2024ophnet}), Cholecystectomy (Cholec80 \cite{twinanda2016endonet}, Cholec80-CVS \cite{rios2023cholec80cvs}, Cholec80-VQA \cite{seenivasan2022surgical}, CholecInstanceSeg \cite{alabi2024cholecinstanceseg}, CholecT50 \cite{Nwoye2022Rendezvous}, Endoscapes2023 \cite{mascagni2024endoscapes2023}), Submucotomy (CoPESD \cite{wang2024copesd}), Myotomy (DSAD \cite{carstens2023dresden}), Nephrectomy (EndoVis2017 \cite{allan20192017}, EndoVis2018 \cite{jimenez2022surgical}, EndoVis2018-VQA \cite{seenivasan2022surgical}, EndoVis2018-VQA-Triplet \cite{seenivasan2022surgical}, LRSP-VQA \cite{du2024lsrp}), Prostatectomy (GraSP \cite{ayobi2023pixel}, PSA-AVA-VQA \cite{seenivasan2022surgical}, SAR-RARP \cite{psychogyios2024sarrarp50}, SurgCoTBench \cite{low2025surgraw}, LRSP-VQA \cite{du2024lsrp}), Gastrojejunostomy (MultiBypass140 \cite{lavanchy2024multibypass140}), Pituitary Surgery (PitVQA \cite{he2024pitvqa}), Colorectal Surgery (SegCol \cite{ju2024segcol}), Lobectomy (SurgCoTBench \cite{low2025surgraw}, LRSP-VQA \cite{du2024lsrp}), Total Mesorectal Excision (LRSP-VQA \cite{du2024lsrp}), Nephroureterectomy (LRSP-VQA \cite{du2024lsrp}), Gastrectomy (LRSP-VQA \cite{du2024lsrp}), Herniorrhaphy (LRSP-VQA \cite{du2024lsrp}), Adrenalectomy (LRSP-VQA \cite{du2024lsrp}), thereby highlighting the breadth of anatomical regions and tasks addressed by current surgical AI benchmarks. The sample rate of each dataset is same with official setting of original dataset, keeping same key frames with annotations. The details of each dataset and how we employ them are shown below.\\

\noindent \textbf{AutoLaparo} The AutoLaparo dataset \cite{wang2022autolaparo} contains 21 laparoscopic hysterectomy videos collected from Prince of Wales Hospital, Hong Kong. The dataset includes annotations for phase recognition, instrument detection, and instrument localization.\\
\textbf{Format}: Open-vocabulary questions\\
\textbf{Image Num (dev/test)}: 56,038/28,705\\
\textbf{Conversation Num (dev/test)}: 57,220/29,577\\
\textbf{Tasks}: Instrument recognition, phase recognition, instrument localization\\\\
\textbf{Cataract-1K} The Cataract-1K dataset \cite{ghamsarian2024cataract} contains 1000 videos of cataract surgeries performed at the Klagenfurt Eye Clinic. Surgery phase annotations are provided for 56 regular videos, and pixel-level annotations of the relevant anatomy and instruments are provided for 2256 frames from 30 cataract surgery videos.\\
\textbf{Format}: Open-vocabulary questions\\
\textbf{Image Num (dev/test)}: 106,985/0\\
\textbf{Conversation Num (dev/test)}: 106,985/0\\
\textbf{Tasks}: Phase recognition\\\\
\textbf{CaDISv2} The CaDISv2 dataset \cite{luengo20222020cataract} contains 4,670 images sampled from 25 cataract surgery videos. Each pixel in these images is labeled with one of 36 classes, including 29 surgical instrument classes, 4 anatomy classes, and 3 miscellaneous object classes.\\
\textbf{Format}: Open-vocabulary questions\\
\textbf{Image Num (dev/test)}: 4,084/586\\
\textbf{Conversation Num (dev/test)}: 15,680/2,436\\
\textbf{Tasks}: Instrument recognition, instrument localization\\\\
\textbf{Cholec80 (SurgVLM-Bench)} The Cholec80 dataset \cite{twinanda2016endonet} contains 80 laparoscopic cholecystectomy videos captured by 13 surgeons at the University Hospital of Strasbourg. Annotations include seven surgical phases (e.g., preparation, gallbladder dissection) and tool presence labels for seven surgical tools (e.g., grasper, clipper).\\
\textbf{Format}: Open-vocabulary questions; Multiple choice questions\\
\textbf{Image Num (dev/test)}: 86,304/98,194\\
\textbf{Conversation Num (dev/test)}: 163,833/182,062\\
\textbf{Tasks}: Instrument recognition, instrument localization\\\\
\textbf{Cholec80-CVS} The Cholec80-CVS dataset \cite{rios2023cholec80cvs} is an extention of the Cholec80 dataset. It provides annotations for all 80 videos, specifically focusing on assessments of the criteria defining the Critical View of Safety (CVS).\\
\textbf{Format}: Multiple choice questions\\
\textbf{Image Num (dev/test)}: 30,550/32,130\\
\textbf{Conversation Num (dev/test)}: 30,550/32,130\\
\textbf{Tasks}: Critical view safety (CVS) assessment\\\\
\textbf{Cholec80-VQA} The Cholec80-VQA dataset \cite{seenivasan2022surgical} contains 21,591 frames sampled from Cholec80 dataset. Based on the tool-operation and phase annotations, two types of question-answer pairs are generated for each frame: (i) Classification-based QnA, where answers are categorized into 14 distinct single-word labels and (ii) Sentence-based QnA, where answers are structured as complete sentences.\\
\textbf{Format}: Multiple choice questions\\
\textbf{Image Num (dev/test)}: 17,043/4,548\\
\textbf{Conversation Num (dev/test)}: 68,172/18,192\\
\textbf{Tasks}: Visual question answering (VQA), with a focus on instrument and tissue recognition.\\\\
\textbf{Cholecinstanceseg} The CholecInstanceSeg dataset \cite{alabi2024cholecinstanceseg} contains 41.9k annotated images extracted from 85 clinical cases and 64.4k tool instances. Each frame is labelled with semantic masks and instance IDs.\\
\textbf{Format}: Open-vocabulary questions\\
\textbf{Image Num (dev/test)}: 28,030/0\\
\textbf{Conversation Num (dev/test)}: 78,953/0\\
\textbf{Tasks}: Instrument recognition, instrument localization\\\\
\textbf{CholecT50 (SurgVLM-Bench)} The CholecT50 dataset \cite{Nwoye2022Rendezvous} contains 50 laparoscopic cholecystectomy videos sampled from Cholec80. It is designed to support advanced surgical activity recognition such as tool-tissue interactions.\\
\textbf{Format}: Open-vocabulary questions; Multiple choice questions\\
\textbf{Image Num (dev/test)}: 70,539/19,263\\
\textbf{Conversation Num (dev/test)}: 70,539/19,263\\
\textbf{Tasks}: Triplet recognition\\\\
\textbf{CoPESD} The CoPESD dataset \cite{wang2024copesd} contains 17,679 images from over 35 hours of ESD videos for both robot-assisted and conventional surgeries. Annotations include target tissues, instruments, surgical motions, motion directions and corresponding bounding boxes.\\
\textbf{Format}: Open-vocabulary questions\\
\textbf{Image Num (dev/test)}: 78,141/10,017\\
\textbf{Conversation Num (dev/test)}: 248,556/31,717\\
\textbf{Tasks}: Triplet recognition\\\\
\textbf{DSAD}: The Dresden Surgical Anatomy Dataset (DSAD) \cite{carstens2023dresden} contains 13,195 laparoscopic images with semantic segmentations of eight abdominal organs, the abdominal wall and two vessel structures in laparoscopic view. Annotations comprise semantic segmentations of single organs and one multi-organ-segmentation dataset including segments for all eleven anatomical structures.\\
\textbf{Format}: Open-vocabulary questions\\
\textbf{Image Num (dev/test)}: 14,625/0\\
\textbf{Conversation Num (dev/test)}: 14,625/0\\
\textbf{Tasks}: Tissue recognition\\\\ 
\textbf{Endoscapes2023 (SurgVLM-Bench)} The Endoscapes2023 dataset \cite{mascagni2024endoscapes2023} contains 201 laparoscopic cholecystectomy videos recorded at the Institute of Image-Guided Surgery (IHU-Strasbourg, France). Annotations include segmentation masks of surgical instruments and hepatocystic anatomy, as well as assessments of the criteria defining the CVS.\\
\textbf{Format}: Multiple choice questions\\
\textbf{Image Num (dev/test)}: 36,694/9,519\\
\textbf{Conversation Num (dev/test)}: 37,873/9,808\\
\textbf{Tasks}: Tissue recognition, critical view safety (CVS) assessment\\\\
\textbf{Endovis2017 (SurgVLM-Bench)} The EndoVis 2017 dataset \cite{allan20192017} contains 10 sequences of abdominal porcine cases recorded using da Vinci Xi systems. The dataset contain 7 different robotic surgical instruments with automatically generated annotations from robot forward kinematics.\\
\textbf{Format}: Open-vocabulary questions\\
\textbf{Image Num (dev/test)}: 1,800/600\\
\textbf{Conversation Num (dev/test)}: 6,117/2,005\\
\textbf{Tasks}: Instrument recognition, instrument localization\\\\
\textbf{Endovis2018} The EndoVis 2018 dataset \cite{jimenez2022surgical} contains 19 sequences from a single porcine training cases recorded on da Vinci X or Xi system using specialized recording hardware. The dataset contain 7 different robotic surgical instruments with segmentation masks of anatomical objects and medical devices.\\
\textbf{Format}: Open-vocabulary questions\\
\textbf{Image Num (dev/test)}: 1,788/1,444\\
\textbf{Conversation Num (dev/test)}: 7,572/6,590\\
\textbf{Tasks}: Instrument recognition, tissue recognition, instrument localization\\\\
\textbf{Endovis2018-VQA (SurgVLM-Bench)} The EndoVis2018-VQA dataset \cite{seenivasan2022surgical} contains 2,007 frames sampled from Endovis2018 dataset. Based on the annotations, two types of question-answer pairs are generated for each frame: (i) Classification-based QnA, where answers are categorized into 26 distinct single-word labels and (ii) Sentence-based QnA, where answers are structured as complete sentences.\\
\textbf{Format}: Open-vocabulary questions; Multiple choice questions\\
\textbf{Image Num (dev/test)}: 1,560/447\\
\textbf{Conversation Num (dev/test)}: 19,588/5,985\\
\textbf{Tasks}: Visual question answering (VQA), with a focus on instrument localiztaion and instrument and tissue recognition.\\\\
\textbf{Endovis2018-VQA-Triplet} The Endovis2018-VQA-Triplet dataset \cite{seenivasan2022surgical} is an extention of the EndoVis2018-VQA dataset. It contains 2,007 images and introduces triplet recognition based on the refined VQA annotations.\\
\textbf{Format}: Open-vocabulary questions\\
\textbf{Image Num (dev/test)}: 1,560/447\\
\textbf{Conversation Num (dev/test)}: 1,560/447\\
\textbf{Tasks}: Triplet recognition\\\\
\textbf{GraSP} The GraSP dataset \cite{ayobi2023pixel} contains 13 robot-assisted radical prostatectomy videos recorded at Fundacion Santa Fe de Bogota. Annotations include 10 surgical phases, 20 step categories, 14 atomic actions and 7 instruments.\\
\textbf{Format}: Open-vocabulary questions\\
\textbf{Image Num (dev/test)}: 73,618/42,897\\
\textbf{Conversation Num (dev/test)}: 161,900/92,641\\
\textbf{Tasks}: Instrument recognition, action recognition, step recognition, phase recognition, instrument localization\\\\
\textbf{MultiBypass140}: The MultiBypass140 dataset \cite{lavanchy2024multibypass140} contains 140 surgical videos of laparoscopic Roux-en-Y gastric bypass (LRYGB) surgeries performed at two medical centers. Annotations include 12 surgical phases and 46 step categories.\\
\textbf{Format}: Open-vocabulary questions\\
\textbf{Image Num(dev/test)}: 539,451/231,166\\
\textbf{Conversation Num (dev/test)}: 1,078,902/462,332\\
\textbf{Tasks}: Phase recognition, step recognition\\\\
\textbf{OphNet}: The OphNet dataset \cite{hu2024ophnet} contains 2,278 surgical videos covering 66 types of cataract, glaucoma, and corneal surgeries. It includes detailed annotations for 102 surgical phases and 150 fine-grained operations, with sequential and hierarchical labeling to enhance interpretability and surgical workflow understanding.\\
\textbf{Format}: Open-vocabulary questions\\
\textbf{Image Num (dev/test)}: 19,278/6,418\\
\textbf{Conversation Num (dev/test)}: 19,278/6,418\\
\textbf{Tasks}: Phase recognition, operation recognition\\\\
\textbf{PitVQA}: The PitVQA dataset \cite{he2024pitvqa} contains 25 endoscopic pituitary surgery videos from the National Hospital of Neurology and Neurosurgery in London, United Kingdom. All videos were annotated with 4 phases, 15 steps, 18 instruments, 3 variations of instruments present in a frame, 5 positions of the instruments, and 14 operation notes in the annotation classes.\\
\textbf{Format}: Open-vocabulary questions\\
\textbf{Image Num (dev/test)}: 84,406/24,767\\
\textbf{Conversation Num (dev/test)}: 680,361/203,881\\
\textbf{Tasks}: Visual question answering (VQA), with a focus on instrument and tissue recognition.\\\\
\textbf{PSA-AVA-VQA} The PSA-AVA-VQA dataset \cite{seenivasan2022surgical} contains 2,244 frames sampled from PSA-AVA dataset. Each frame includes both sentence-form and single-word classification answers. The conversations are further categorized into location-based, phase-based, and step-based types, providing structured learning for surgical AI models.\\
\textbf{Format}: Open-vocabulary questions\\
\textbf{Image Num(dev/test)}: 2,244/0\\
\textbf{Conversation Num (dev/test)}: 16,095/0\\
\textbf{Tasks}: Visual question answering (VQA), with a focus on action, step and phase recognition\\\\
\textbf{SAR-RARP (SurgVLM-Bench)} The SAR-RARP dataset \cite{psychogyios2024sarrarp50} contains 50 suturing video segments of Robotic Assisted Radical Prostatectomy (RARP). The dataset annotates nine semantic classes labeling all areas of the surgical scene that correspond to surgical instrumentation.\\
\textbf{Format}: Open-vocabulary questions; Multiple choice questions\\
\textbf{Image Num(dev/test)}: 130,232/32,473\\
\textbf{Conversation Num (dev/test)}: 130,232/32,473\\
\textbf{Tasks}: Action recognition\\\\
\textbf{SegCol}: The SegCol dataset \cite{ju2024segcol} contains 2,560 images sampled from different real colonoscopy videos (EndoMapper dataset). Images are labelled with pixel-level masks of 3 instrument classes and 1 class representing the edges of colon folds. \\
\textbf{Format}: Open-vocabulary questions\\
\textbf{Image Samples (dev/test)}: 2,560/0\\
\textbf{Image Conversation Num (dev/test)}: 3,755/0\\
\textbf{Tasks}: Instrument recognition, instrument localization\\\\
\textbf{SurgCoTBench} The SurgCoTBench dataset \cite{low2025surgraw} is a reasoning-based benchmark for robotic-assisted surgery, built from 12 patients undergoing prostatectomy and lobectomy. Each frame is annotated with up to seven questions covering five types: instrument recognition, action recognition, action prediction, patient data extraction, and outcome assessment. Although the dataset was originally designed to support reasoning tasks, we do not apply chain-of-thought (CoT) prompting or reasoning-style prompts in our experiments.\\
\textbf{Format}: Multiple choice questions\\
\textbf{Image Num (dev/test)}: 2,277/0\\
\textbf{Conversation Num (dev/test)}: 14,176 /0\\
\textbf{Tasks}: Visual question answering (VQA), with a focus on instrument and action recognition and action prediction.

\section{Benchmark Construction}
From SurgVLM-DB, we selected six datasets to build SurgVLM-Bench, ensuring coverage of the full spectrum of surgical AI tasks. We include EndoVis2017 for instrument localization and EndoVis2018 VQA for vision-grounded question answering; Cholec80 for phase recognition, CholecT50 for compositional triplet reasoning, and SAR-RARP for fine-grained action recognition; and Endoscape2023 for critical-view-of-safety assessment. Together, these six benchmarks span various tasks requiring visual perception, temporal analysis, to the high-level reasoning capability. 
Regarding the sample amount, given that large cost burdens when evaluating using closed-sourced commercial APIs, we randomly sample 1000 frames from the official test split in each dataset, and do the comparison leaderboard. We evaluate all open-source VLMs with same hyper-parameters (e.g. temperature, top\_p, top\_k) and system prompts under same VLLM version and GPU environments, can reduce randomness and yield consistency in multiple runs on the small data subset. Thus, the current benchmark can well reflect the superiority of our method.

\subsection{Evaluation Metrics}

Each SurgX-MMBench task uses established, task-appropriate metrics. Below we provide intuitive explanations alongside formal definitions, so that readers from all backgrounds can appreciate what each metric measures.

\subsubsection{Instrument Localization (EndoVis2017)}  
In the instrument localization task, models must draw bounding boxes around up to ten instrument types. We use the following metrics to evaluate model performance:  \\
\textbf{Mean IoU (mIoU)}: for each predicted box, the Intersection-over-Union with the matching ground‐truth box; averaged across all boxes. Higher means tighter, more precise boxes.\\
\textbf{mAP@50 / mAP@75}: average precision of detections at IoU thresholds of 0.5 and 0.75, indicating how well the model localizes instruments to within moderate or strict overlap. \\
\textbf{COCO AP}: the average precision computed over multiple IoU thresholds from 0.5 to 0.95 in steps of 0.05, giving a single summary of localization quality.

\subsubsection{Instrument and Tissue Recognition (EndoVis2018-VQA)}  
In the instrument and tissue recognition tasks, models are provided with image–question pairs and are asked to provide open vocabulary answers. We measure:  \\
\textbf{Answer Accuracy}: the percentage of exact matches between the model’s answer and the ground truth keyword, emphasizing the need for precise surgical terminology.  \\
\textbf{BLEU-4}: compares 4-gram overlaps between the candidate answer and reference—higher scores indicate more shared phrases.  \\
\textbf{METEOR}: combines unigram precision and recall, with stemming and synonym matching, to reward semantically correct answers even if phrased differently.  \\
\textbf{ROUGE-1}: measures the recall of unigrams (single words) from the reference in the candidate, highlighting how much key content the model captures.

\subsubsection{Phase Recognition (Cholec80)} 
In the phase recognition task, models must assign each image to one of seven predefined surgical phases, selecting the label that most accurately reflects the procedure’s current step. The evaluation metrics are as follows:
\textbf{Accuracy}: the percentage of frames whose predicted phase label exactly matches the ground truth. It reflects overall correctness but can be skewed if one phase is much more frequent.  \\
\textbf{Recall}: measures the proportion of true positive frames that the model correctly identifies. A high recall means the model rarely misses that phase. Formally:  
\[
\text{Recall} = \frac{\text{True Positives}}{\text{True Positives} + \text{False Negatives}}\,.  
\]  
\textbf{Precision}: measures the proportion of frames the model labels as that phase are actually correct. High precision indicates few false alarms:  
\[
\text{Precision} = \frac{\text{True Positives}}{\text{True Positives} + \text{False Positives}}\,.  
\]  
\textbf{Jaccard Index (Intersection-over-Union)}: quantifies overlap between predicted and ground-truth frame sets for each phase. It penalizes both missed frames and false detections:  
\[
\text{Jaccard} = \frac{\text{True Positives}}{\text{True Positives} + \text{False Positives} + \text{False Negatives}}.  
\]

\subsubsection{Action Recognition (SAR-RARP)}  
In the action recognition task, models must assign each image to one of eight classes (seven bimanual surgical gestures or a background class). Performance is measured by accuracy, recall, precision and Jaccard index, using the same evaluation protocol as for phase recognition.

\subsubsection{Triplet Recognition (CholecT50)}  
In the triplet recognition task, each image is annotated with a three part label of instrument, verb and target. Models must correctly identify the instrument, the verb and the target from the available options. The evaluation metrics are:\\
\textbf{Component Accuracy:} the fraction of frames for which the instrument, verb, or target is correctly identified.\\
\textbf{Triplet Accuracy:} the fraction where all three components are correct simultaneously—this is the strictest measure of full surgical intent understanding. \\
\textbf{Mean Average Precision (mAP):} for each component and for the full triplet, we rank model confidence scores and compute the area under the precision–recall curve. Averaging these areas yields mAP, reflecting both ranking quality and detection performance.

\subsubsection{Critical View of Safety (CVS) Evaluation (Endoscape2023 CVS)}  
In the CVS evaluation task, each frame is labeled for three binary Critical View of Safety checks: cystic plate exposure, lower gallbladder clearance and two structure identification. We compute two accuracy metrics to assess performance:\\
\textbf{Average Accuracy}: the overall fraction of correct yes/no predictions across all three criteria. \\
\textbf{Balanced Accuracy}: for each criterion, the average of sensitivity and specificity (to account for class imbalance between “present” vs. “absent”), then averaged across the three criteria.

\end{document}